\documentclass[sigconf,screen,nonacm]{acmart}

\setcopyright{none}
\renewcommand\footnotetextcopyrightpermission[1]{}

\usepackage{booktabs}
\usepackage{tcolorbox}
\tcbuselibrary{breakable}
\usepackage{multirow}
\usepackage[table]{xcolor}
\usepackage{graphicx}
\usepackage{fontawesome5}

\begin{document}

\title{CameraEditor: Camera-Controlled Image Editing via Video-Prior Sequential Modeling}

\author{\texorpdfstring{\begin{tabular}{c}
\normalsize Xin Shen$^{1*}$ \qquad Chengyou Jia$^{1*}$ \qquad Keshuo Xing$^1$
\qquad Zifeng Zhu$^1$ \qquad Changliang Xia$^1$\\[2pt]
\normalsize Bowen Ping$^1$ \qquad Zhuohang Dang$^1$ \qquad Hangwei Qian$^2$
\qquad Minnan Luo$^{1\dagger}$
\end{tabular}}{Xin Shen; Chengyou Jia; Keshuo Xing; Zifeng Zhu; Changliang Xia;
Bowen Ping; Zhuohang Dang; Hangwei Qian; Minnan Luo}}
\affiliation{%
  \institution{\normalfont\small
  $^1$Xi'an Jiaotong University, China \qquad
  $^2$Agency for Science, Technology and Research (A*STAR), Singapore\\[1pt]
  \href{https://github.com/King-Wood-Shen/CameraEditor}{\faGithub\enspace Code}
  \qquad
  \href{https://huggingface.co/keenwood/CameraDiT}{%
    \raisebox{-0.18em}{\includegraphics[height=1.15em]{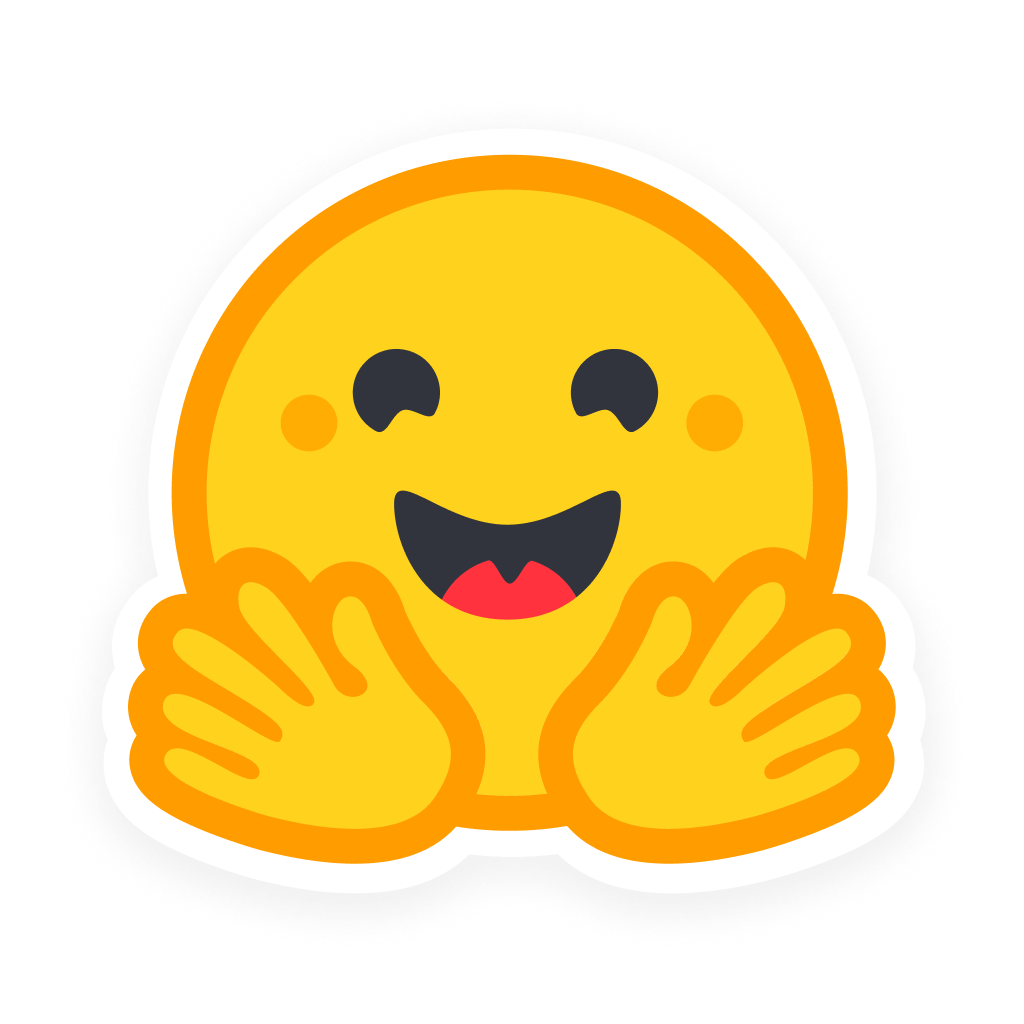}}\enspace Model}
  \quad
  \href{https://huggingface.co/datasets/keenwood/CamEditor-Bench}{Data}
  \qquad
  \href{https://king-wood-shen.github.io/CameraEditor-Home/}{%
    \faHome\enspace Homepage}\\[1pt]
  \footnotesize $^*$Equal contribution \qquad $^\dagger$Corresponding author}
  \city{}
  \country{}}

\renewcommand{\shortauthors}{Shen et al.}

\begin{abstract}
Beyond semantic content, camera parameters play a pivotal role in dictating the geometric perspective and appearance of any given image. While recent image editing models excel at semantic and stylistic manipulation, they struggle with explicit camera parameter control. When handling large perspective shifts, instruction-driven models face a dilemma: they either suffer from structural tearing or generate conservative outputs that ignore geometric instructions. To address this, we introduce CameraEditor, a framework that reformulates camera-controlled editing from a spatial problem into a temporal sequence prediction task. By leveraging the temporal coherence of video diffusion models, our approach integrates an explicit geometric perception module with a dynamic reference routing mechanism. This allows us to construct geometrically rigorous visual reference pairs via dynamic panorama cropping, overcoming the ambiguity of text-based instructions. Furthermore, CameraEditor strategically inserts intermediate transition frames to decompose large perspective shifts, providing a robust temporal buffer that preserves content identity and spatial coherence. We construct a training dataset of 5,760 instances. As an independent contribution, we introduce CamEditor-Bench, a model-agnostic evaluation suite of 462 test cases. Extensive experiments demonstrate that CameraEditor achieves state-of-the-art camera control precision and source identity preservation, outperforming existing methods.
\end{abstract}


\begin{teaserfigure}
  \centering
  \includegraphics[width=0.9\textwidth]{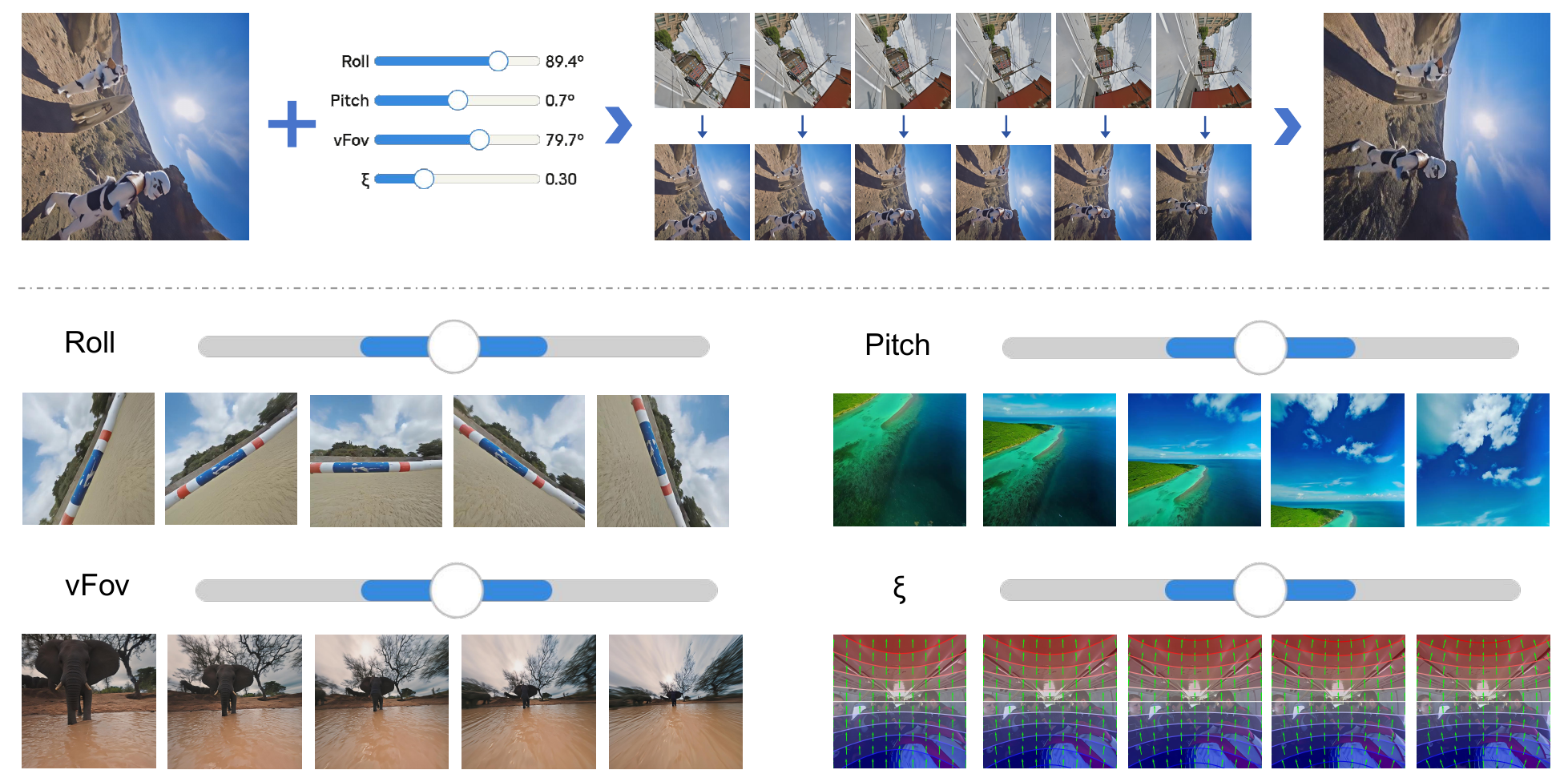}
  \caption{Overview of CameraEditor. We introduce a video-prior framework for precise camera-parameterized editing. (Top) Large perspective shifts are decomposed via intermediate frame interpolation. (Bottom) CameraEditor achieves state-of-the-art precision across diverse scenes, preserving fine-grained details even under extreme rotations and lens distortions.}
  \Description{A schematic overview and qualitative examples of CameraEditor for camera-parameterized image editing.}
  \label{fig:teaser}
\end{teaserfigure}

\maketitle


\section{Introduction}
\label{sec:intro}

\begin{figure*}[htbp]
    \centering
    \includegraphics[width=0.9\textwidth]{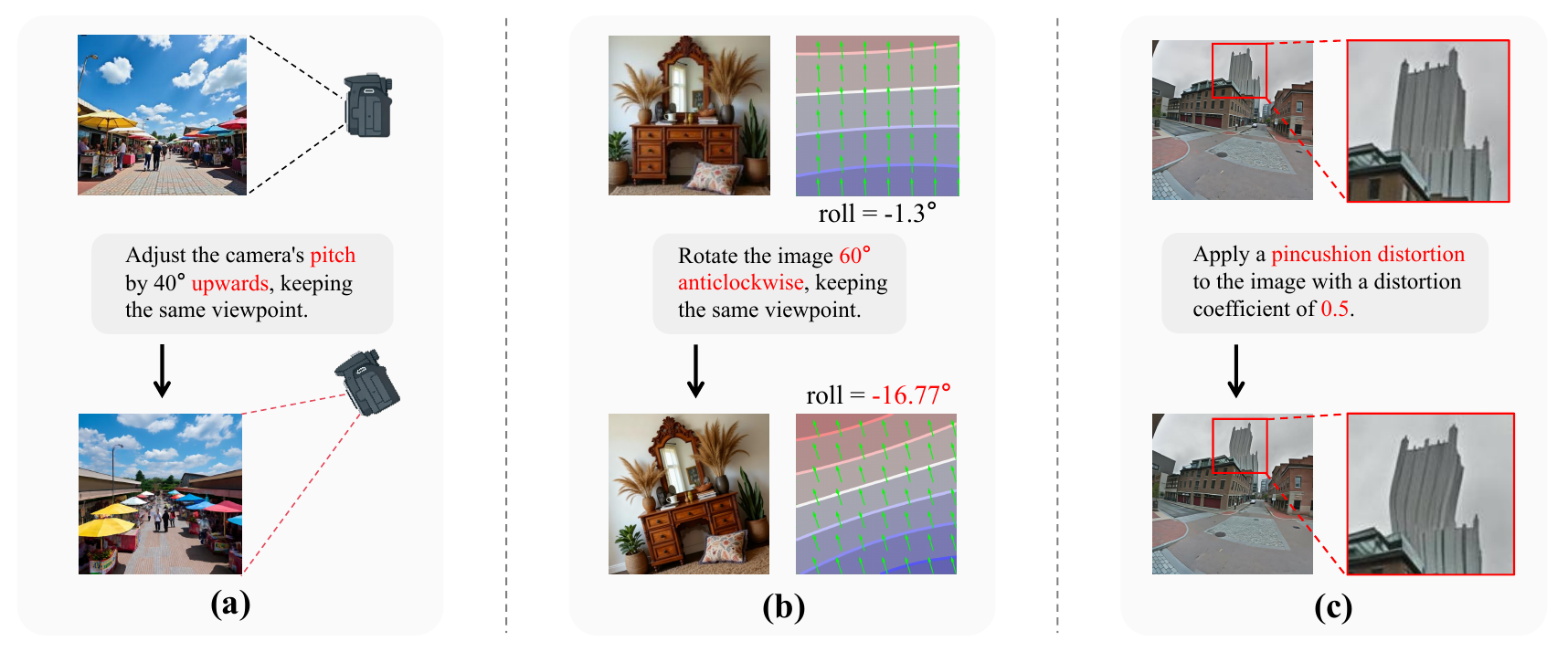}
    \caption{\textbf{Limitations of instruction-driven models.} Ambiguous text prompts without mathematical constraints cause three critical failures: \textbf{(a) Terminology Misinterpretation} (conflating geometric shifts with semantic changes), \textbf{(b) Parametric Imprecision} (failing exact-degree adjustments), and \textbf{(c) Content Identity Loss} (structural tearing under complex distortions).}
    \Description{Examples of terminology errors, inaccurate camera changes, and identity loss produced by instruction-driven editing models.}
    \label{fig:limitation}
\end{figure*}

Recently, image editing has witnessed remarkable progress \cite{zhang2025unified}. Not only closed-source models such as GPT-Image-1.5 \cite{gptimage15} and Nano-Banana-Pro \cite{nanobananapro}, but also some open source models like Qwen-Image-Editing \cite{wu2025qwen} and Flux.2 \cite{flux2dev} have demonstrated impressive capabilities in manipulating images based on user instructions. They can perform various editing tasks \cite{huang2025diffusion}, including semantic editing \cite{rout2024semantic,yu2025anyedit,huang2025pfb}, stylistic editing \cite{han2025stylebooth, lei2025stylestudio}, and spatial editing \cite{ yan2025eedit}. However, an image's composition is shaped by both its content and camera parameters (\emph{e.g.}, focal length, aperture, pose). Even when keeping the content unchanged, adjusting camera parameters can significantly alter the visual impression. In that case, editing images through camera parameters is essential for precise and controlled visual manipulation. Unfortunately, they often fall short in delivering accurate and fine-grained edits aligned with specific camera parameters adjustments.

As demonstrated in Figure \ref{fig:limitation}, these limitations manifest in three critical dimensions:
(1) Terminology Misinterpretation: General-purpose models lack rigorous geometric understanding, often conflating precise camera parameters (\emph{e.g.}, ``pitch'' with broad semantic concepts like ``aerial view''). (2) Parametric Imprecision: Operating without exact mathematical anchors, existing methods struggle to execute fine-grained rotation or exact-degree geometric adjustments (\emph{e.g.}, failing to reach the instructed $60^{\circ}$ rotation). (3) Identity Preservation Conflict: Under large perspective shifts, models face a dilemma: they either suffer structural tearing (\emph{e.g.}, the severely warped building) to follow instructions, or generate conservative outputs that ignore geometric changes to preserve identity.

To address these limitations, we formulate the task as editing a source image under specified camera-parameter transformations and propose a framework for camera-parameterized image editing, named CameraEditor. Unlike conventional methods that rely on pre-trained text-to-image models for image editing \cite{jin2023perspective, he2024cameractrl}, our approach reformulates this editing task as a sequence prediction problem, which helps the model fully leverage video priors. Furthermore, it brings two additional benefits: (1) By leveraging video models, we use a reference image pair consisting of a source image and its edited version to guide the model in applying consistent camera-parameter transformations to the target image. This approach overcomes the limitations of text-based prompts, which often struggle to capture precise geometric changes, and pixel-level control maps that may distort overall structure. (2) Inspired by the Chain of Frames (CoF) \cite{ghazanfari2025chain} approach, our method inserts transition frames for both reference and target image pairs. By aligning these transition steps frame-by-frame, we facilitate highly smooth, consistent transformations. We effectively decompose large-magnitude geometric transformations into a traceable trajectory, thereby ensuring smooth and consistent results.

To support our framework, we construct a dataset generation pipeline that synthesizes a training dataset containing 5,760 instances with detailed camera parameter annotations from both real-world images and synthetic scenes generated with Unreal Engine 5 (UE5) \cite{el2022unreal}. Independently of the proposed model, we introduce CamEditor-Bench, a model-agnostic evaluation suite consisting of 462 strictly isolated test instances. Crucially, alongside this benchmark, we pioneer a novel Camera Alignment evaluation protocol. By leveraging Perspective Field representations, this metric system effectively avoids traditional estimation noise, allowing for an explicit and rigorous assessment of absolute geometric precision.
Extensive experiments on CamEditor-Bench demonstrate that CameraEditor achieves state-of-the-art performance in both geometric alignment and content preservation. By reformulating camera control as a temporal video generation task, our visual reference pairs eliminate text-based misinterpretations, while the frame-by-frame decomposition effectively buffers extreme pixel displacements to prevent structural collapse.
In summary, CameraEditor and CamEditor-Bench constitute two independent contributions, supported by comprehensive evaluation:
\begin{itemize}
\item We present \textbf{CameraEditor}, which reformulates camera-controlled editing into a temporal sequence prediction task. By integrating dynamic reference routing to construct rigorous visual prompts and inserting intermediate transition frames, it circumvents the inherent ambiguity of text-based instruction control and buffers massive perspective shifts to effectively prevent structural collapse.

\item We construct a 5,760-instance training dataset for CameraEditor. Separately, we introduce \textbf{CamEditor-Bench}, a model-agnostic evaluation suite of 462 perspective-image pairs derived from diverse real-world and UE5 panoramas, complemented by an optimization-free geometric evaluation protocol based on Perspective Fields.

\item Automatic and human evaluations with same-backbone controls show that the gains arise from explicit geometric conditioning and sequential visual references rather than backbone capacity alone.

\end{itemize}


\section{Related Work}
\label{sec:related_work}

\subsection{Image Editing}
Image editing has advanced with instruction-driven approaches
\cite{brooks2023instructpix2pix}, where models are fine-tuned on large
``instruction-response'' datasets for intuitive manipulation
\cite{betker2023improving}. These methods evolved from fine-tuning diffusion
models \cite{rombach2022high} to flexible editing techniques that exploit
pre-trained generative priors \cite{xu2023cyclenet, ruiz2023dreambooth}.
Recent training-free approaches further avoid task-specific adaptation. For
instance, GIDE \cite{zhu2026gide} combines grounding with discrete noise
inversion for diffusion large language models, enabling localized edits while
preserving unedited regions. Nevertheless, such methods primarily target
semantic or region-based content manipulation rather than explicit
camera-parameterized geometric transformations. Instruction-driven editors
have also refined datasets and optimized architectures
\cite{zhang2025context} to enhance control and realism. Recent advances improve
task efficiency through dataset and model optimizations. Lightweight methods
like LoRA \cite{huang2024context} enable task-specific adaptation without large
retraining, though task parameters remain necessary. Furthermore, the
integration of novel proprietary systems within multimodal LLMs
(\emph{e.g.}, GPT-4o \cite{hurst2024gpt} and Gemini
\cite{imran2024google}) has further blurred the boundary between
conversational dialog and visual editing, while commercial platforms like
Midjourney \cite{tsidylo2023artificial} and RunwayML
\cite{esser2023structure} embed these models into end-to-end creative
workflows.

Beyond pure text or instruction guidance, precise image manipulation fundamentally requires robust editing under diverse structural conditions. To resolve the spatial ambiguity of natural language, conditional frameworks incorporate auxiliary modalities to rigidly constrain the diffusion process. Methods like ControlNet  \cite{zhang2023adding} and T2I-Adapter \cite{mou2024t2i} utilize depth maps, semantic segmentations, and Canny edges to dictate fine-grained geometric structures, whereas layout-guided models \cite{li2023gligen} enforce strict bounding-box object placement to maintain compositional integrity. However, despite these significant strides in handling diverse 2D conditional inputs, existing architectures often struggle when the condition involves explicit extrinsic camera parameters or 3D-aware novel view synthesis \cite{qin2025camedit}. While planar conditions successfully dictate 2D layouts, they inherently lack the underlying geometric priors required to maintain view-dependent structural consistency and object permanence during complex camera trajectory transformations, exposing a critical architectural void.
However, current models struggle with fine-grained edits requiring precise camera control, such as rotation angles and Perspective Field, limiting their use in tasks like cinematography.

\subsection{Visual Camera Geometry}
Camera parameter estimation and geometric control are foundational to 3D computer vision  \cite{pollefeys1999self}. Classical methods rely on multi-view constraints and projective geometry for rigorous calibration and structure-from-motion  \cite{hartley2003multiple, schonberger2016structure}. Deep learning later shifted toward directly regressing camera poses from raw pixels  \cite{kendall2015posenet, hold2018perceptual, workman2015deepfocal}. However, mapping high-dimensional images to rigid $SE(3)$ manifolds is highly non-linear, making direct regression ill-posed and prone to poor generalization  \cite{sinha2023sparsepose}.

To bridge parameter estimation and visual synthesis, methodologies introduced intermediate representations like displacement fields  \cite{liao2020model, zhuang2021fusing}. Concurrently, implicit neural representations  \cite{mildenhall2021nerf} and diffusion-based novel view synthesis (NVS)  \cite{liu2023zero, shi2023zero123++} manipulated viewpoints using relative coordinates. Yet, these NVS frameworks are predominantly object-centric, failing to execute precise adjustments of camera extrinsics and intrinsics in complex, open-domain scenes, which severely limits their practical applicability.

Recently, controllable diffusion models incorporated explicit camera viewpoint modeling. Frameworks like CameraCtrl  \cite{he2024cameractrl} and MotionCtrl  \cite{wang2024motionctrl} inject camera matrices to dictate trajectories during generation, while others explore spatial conditioning for text-to-image synthesis  \cite{zheng2024cami2v}. Despite these advancements, their applicability is strictly restricted to the forward generation phase. For instance, while PreciseCam  \cite{bernal2025precisecam} achieves accurate camera control during generation, it lacks an inversion mechanism to re-render the latent geometry of an existing input. Consequently, manipulating exact camera parameters on pre-existing images without inducing severe structural degradation remains an unresolved bottleneck in post-capture editing scenarios.


\section{Methodology}

\begin{figure*}[t]
    \centering
    \includegraphics[width=0.85\textwidth]{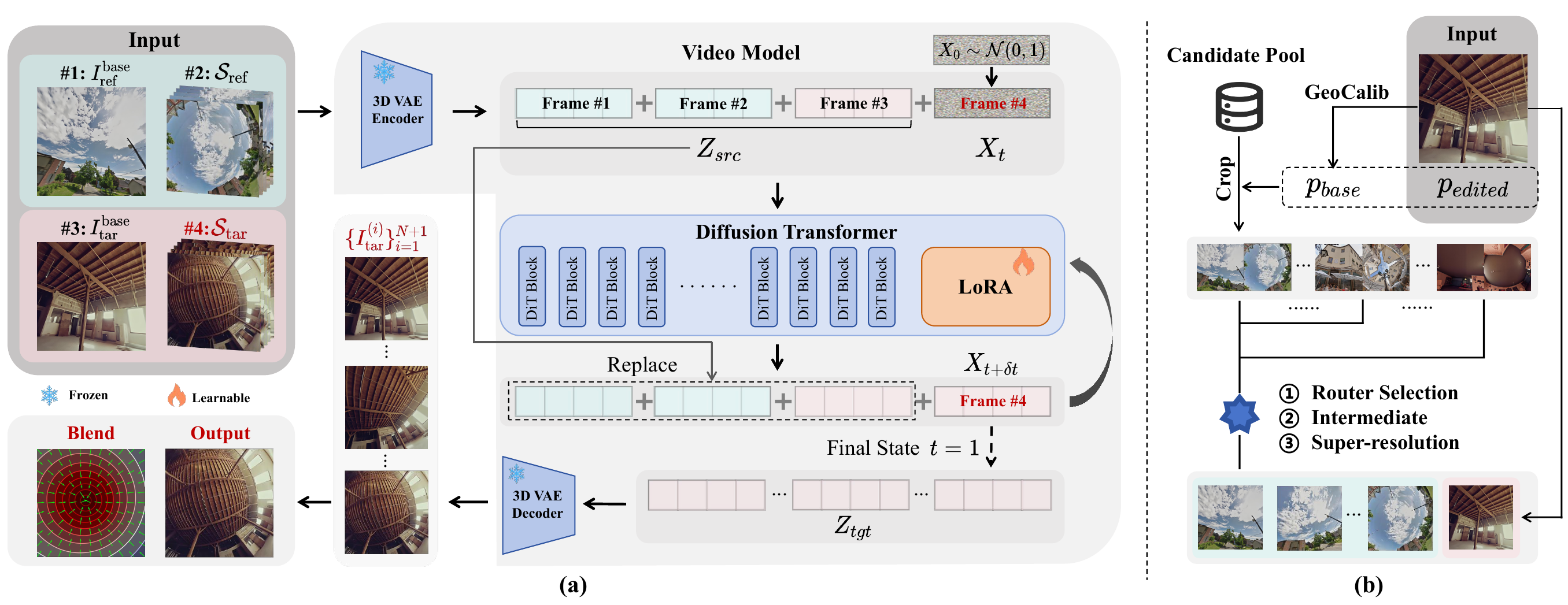}
    \caption{Overview of CameraEditor. (a) Generation Pipeline: A video diffusion model generates the target Chain of Frames (CoF) sequence conditioned on the source image and reference visual sequence. (b) Dynamic Routing: A routing mechanism selects the optimal reference prior via GeoCalib during inference for precise viewpoint alignment.}
    \Description{CameraEditor architecture showing the video diffusion generation pipeline and GeoCalib-based dynamic reference routing.}
    \label{fig:Architecture}
\end{figure*}

This section details CameraEditor, a framework that reformulates camera-controlled image editing into a sequence prediction task (Figure \ref{fig:Architecture}). The generation process is driven by a video generative backbone (Sec. \ref{sec:3.1}) and conditioned on explicit reference visual sequences (Sec. \ref{sec:3.2}). To effectively bridge large perspective shifts, we introduce a Chain of Frames (CoF) strategy to decompose complex geometric transformations (Sec. \ref{sec:3.3}). Finally, a dynamic routing mechanism is employed during inference to ensure strict viewpoint alignment without retraining (Sec. \ref{sec:3.4}).

\subsection{Task Definition and Scope}
\label{sec:task_definition}

We represent a camera configuration as
\[
p=(\mathrm{yaw},\mathrm{roll},\mathrm{pitch},\mathrm{vFoV},\xi),
\]
where $\xi$ denotes radial distortion. Given a source image
$I_{\mathrm{src}}$ with configuration $p_{\mathrm{base}}$ and a bounded target
configuration $p_{\mathrm{edited}}$, the goal is to generate
$I_{\mathrm{edited}}$ whose projected geometry follows
$p_{\mathrm{edited}}$ while preserving the underlying scene content and
identity. CameraEditor controls roll, pitch, vFoV, and radial distortion while
holding yaw fixed. The projected image layout changes according to the target
camera, but the edit does not request a new scene or a different subject.
Constraining the parameter shift keeps the task within camera-parameterized
editing rather than unconstrained outpainting; the exact sampling bounds are
given in the Supplementary Material.

\subsection{Video Generative Models for Image Editing}
\label{sec:3.1}

Video generative models have shown promise in image editing tasks by formulating them as frame-to-frame generation problems  \cite{kong2024hunyuanvideo, lu2023vdt, ma2024latte}. In this paradigm, given a source image $I_{src}$ and desired modifications, the model generates an edited image $I_{tgt}$ while maintaining content consistency across the frames. Specifically, the model focuses on ensuring that the high-level structure and semantics of the original image remain intact, while modifying the details according to the desired target. To achieve this, the model encodes both the source and target images into a shared latent space via a VAE encoder  \cite{tomczak2018vae}, which results in the latent representations $Z_{src}$ and $Z_{tgt}$, respectively.

During inference, starting from pure noise $X_0 \sim \mathcal{N}(0, 1)$, noise is injected at each step to form $X_t$, which is then concatenated with the source latent as $S_t = [Z_{src}, X_t]$. The model $v_\theta$ iteratively denoises this sequence to predict the clean latent $X_1$ of the target frame. At each step, the condition token may become slightly polluted, represented as $Z_{src}^{t+\delta t}$, requiring replacement with the original $Z_{src}$ to ensure consistency and preserve the source image content.

During training, we utilize a flow-matching loss \cite{lipman2022flow} to optimize our model. Specifically, we define the clean target latent as the data endpoint, $X_1 \equiv Z_{tgt}$, and sample pure noise $X_0 \sim \mathcal{N}(0, 1)$. The intermediate noisy state $X_t$ is constructed via linear interpolation:
\begin{align*}
    X_t = (1 - t) X_0 + t X_1,
\end{align*}
where the timestep $t$ is uniformly sampled from $U(0, 1)$. The model is trained to predict the velocity field $V_t = \frac{dX_t}{dt} = X_1 - X_0$, which governs the trajectory towards the clean latent, allowing the estimation of $X_{t+\delta t}$ during inference. The optimization objective is formulated as:
\begin{align*}
     \mathcal{L} = \mathbb{E}_{t, X_{0}, X_{1}} \left[ \| v_\theta(S_t, t_{tgt}) - V_t \|_2^2 \right],
\end{align*}
where $v_\theta(S_t, t)_{tgt}$ denotes the predicted velocity for the target frame.

\subsection{Reference Visual Sequence as Prompt}
\label{sec:3.2}

Prior work typically implemented camera-controlled editing by injecting control signals such as textual descriptions of camera parameters or task-specific attributes like PF-US maps into the model. While these methods offer a certain level of control, they are often hindered by inherent limitations. For instance, textual descriptions often suffer from misalignment between language and scene geometry, making it difficult for the model to accurately translate abstract language into precise geometric transformations. Although PF-US maps effectively control pixel-level details, they fail to maintain the image's overall structure, often causing distortions such as warping or misalignment that compromise spatial coherence and visual integrity.

To overcome these ambiguities, we leverage the temporal attention of video generative models by introducing a \textit{reference visual sequence} as a geometric prompt. Instead of relying on high-level abstract text descriptions, we provide a reference pair $(I_{\text{ref}}^{\text{base}}, I_{\text{ref}}^{\text{edited}})$ that explicitly demonstrates the desired camera parameter variation. Crucially, this reference pair undergoes the exact same camera configuration shift ($p_{\text{base}} \rightarrow p_{\text{edited}}$) as the user's target image pair $(I_{\text{tar}}^{\text{base}}, I_{\text{tar}}^{\text{edited}})$. By concatenating the reference sequence ahead of the target sequence along the temporal dimension, the model's temporal attention mechanisms first internalize the geometric shift from the reference prior. Guided by this strict temporal prompt, the model seamlessly transfers the learned transformation to synthesize the exact $I_{\text{tar}}^{\text{edited}}$ from the input $I_{\text{tar}}^{\text{base}}$, effectively bypassing the spatial ambiguities of traditional text-based editing instructions.

\subsection{CoF-based Intermediate Frames}
\label{sec:3.3}

Large camera-parameter shifts ($p_{\text{base}} \rightarrow p_{\text{edited}}$) cause geometric distortions and content inconsistencies when generating targets directly from the source. Diffusion models struggle to simultaneously resolve complex spatial transformations and preserve scene identity in a single generative step.

To address this, we introduce a Chain of Frames (CoF) strategy that decomposes the large camera transformation into a sequence of manageable steps for both the reference and target domains. Formally, instead of directly predicting the final edited target from the source $I_{\text{tar}}^{base}$, we formulate the generation as predicting a temporally coherent sequence $\mathcal{S}_{\text{tar}} = \{I_{\text{tar}}^{(i)}\}_{i=1}^{N+1}$, where $I_{\text{tar}}^{(N+1)} \equiv I_{\text{tar}}^{\text{edited}}$. Notably, this progression strictly aligns with the reference sequence $\mathcal{S}_{\text{ref}}$. We uniformly interpolate the camera parameters, assigning shared intermediate poses $p^{(i)}$ to both $I_{\text{ref}}^{(i)}$ and $I_{\text{tar}}^{(i)}$ at each step $i$.

The video diffusion model's temporal modeling capabilities are well-suited for processing these frame sequences. The 3D attention mechanisms capture correlations across frames, learning to predict each frame by attending to temporal neighbors and conditioning information. This temporal coherence, combined with gradual camera-parameter changes, enables smooth and consistent transformations while preserving scene content.

\subsection{Inference-Time Viewpoint Alignment}
\label{sec:3.4}

While the generation mechanisms detailed above establish our core framework, a critical challenge remains: during inference, the model must accurately align the input image's viewpoint with the reference pair to ensure that the learned transformations are correctly applied. To achieve this, we implement a two-stage alignment process (Figure~\ref{fig:Architecture}(b)). First, we utilize a geometric calibration module (GeoCalib \cite{veicht2024geocalib}) to estimate the camera parameters of the input image. This module is trained to predict the camera parameters directly from the image, providing a mathematically precise viewpoint estimation. Second, we introduce a Dynamic Reference Routing mechanism to obtain a high-quality reference image that closely aligns with the estimated camera configuration. We dynamically crop a continuous reference sequence from a pre-defined candidate pool based on the estimated and target parameters. A Vision Language Model scores each candidate along five axes---geometric correlation, information balance, scene richness, image clarity, and content coherence---and the minimum axis score determines its overall score. We retain the highest-ranked candidate, ensuring that a weak criterion cannot be hidden by strong scores on the remaining axes. The resulting reference sequence demonstrates the requested transformation without sharing scene identity with the user input. Details of the candidate pools, cropping procedure, and rating prompt are provided in Section \ref{sec:4.1} and the Supplementary Material.


\section{Dataset and Evaluation Benchmark}

\begin{figure*}[htbp]
    \centering
    \includegraphics[width=0.85\textwidth]{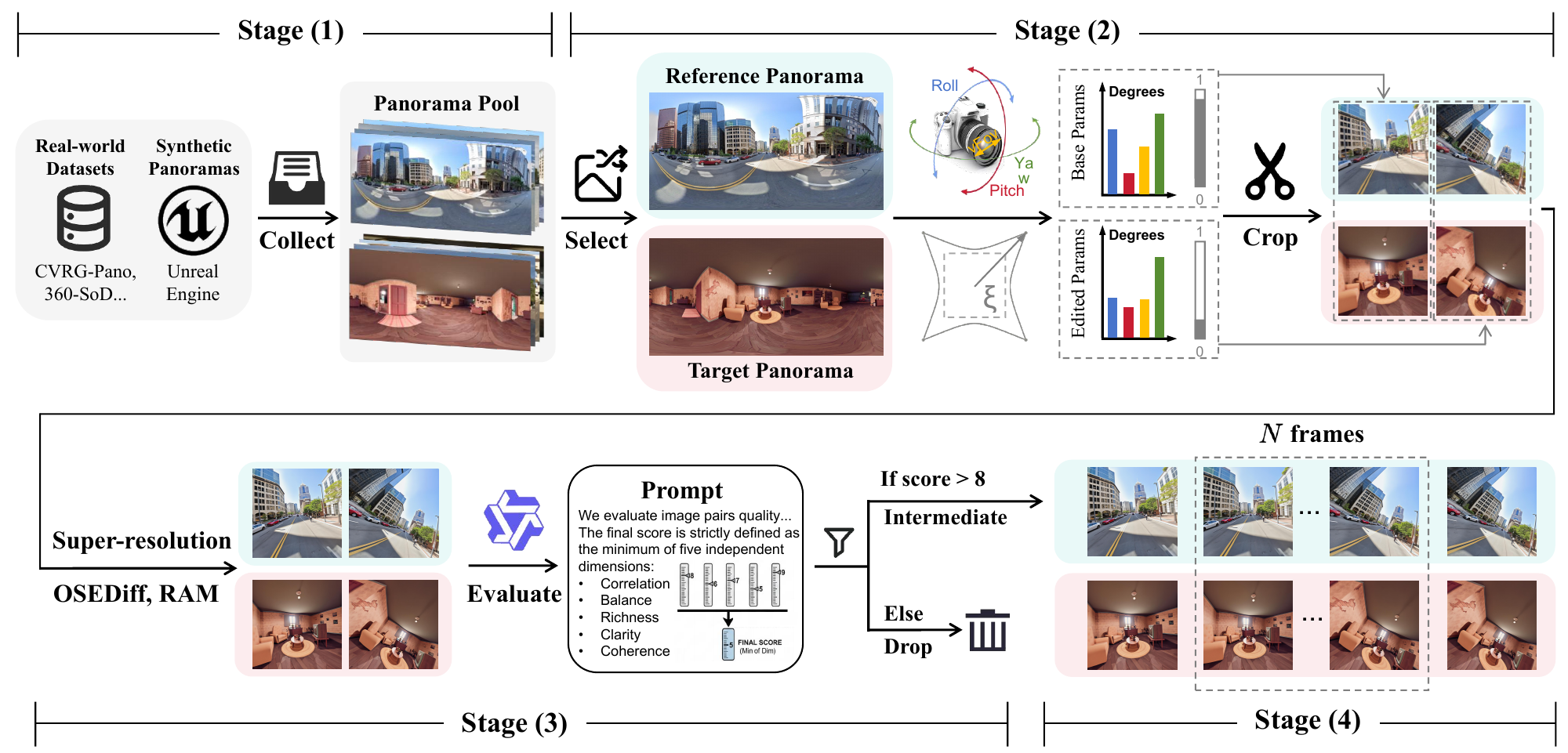}
    \caption{The dataset generation pipeline. Panoramas are parameterized and cropped into reference-target pairs. Following super-resolution and VLM filtering, intermediate frames are generated to form continuous temporal sequences.}
    \Description{Dataset construction pipeline from panoramic images through cropping, filtering, and temporal sequence generation.}
    \label{fig:dataset_pipeline}
\end{figure*}

This section details our data generation and benchmarking protocol. We first construct a specialized training dataset of 5,760 panoramic instances (Sec. \ref{sec:4.1}). Next, we introduce CamEditor-Bench for rigorous zero-shot evaluation (Sec. \ref{sec:4.2}), and finally define robust, optimization-free geometric metrics (Sec. \ref{sec:4.3}).

\subsection{Training Dataset Construction}
\label{sec:4.1}

Since sequential data with continuous camera transitions is absent in current editing domains, we construct an automated pipeline (Figure \ref{fig:dataset_pipeline}) to generate 5,760 training instances in four stages.

(1) \textbf{Various panoramic images collection:} We collect high-quality panoramic images from real-world datasets such as 360-SoD \cite{li2019distortion}, F360-SoD \cite{orhan2022semantic} and CVRG-Pano \cite{zhang2020fixation}, and synthetic scenes rendered using Unreal Engine 5 (UE5). This strategic combination of real-world photorealism and precisely controlled synthetic environments ensures our compact dataset covers a comprehensive blend of diverse geometric structures across indoor and outdoor settings, ultimately establishing a panorama pool for subsequent selection.


(2) \textbf{Image Selection \& Parameter Cropping:} To construct each training instance, we first select a reference panorama and a target panorama from our panorama pool. We then model cameras via Perspective Fields (yaw, pitch, roll, vFov, and radial distortion $\xi$). Edited parameters $p_{\text{edited}}$ are randomly sampled, while base parameters $p_{\text{base}}$ are tightly constrained to prevent unconstrained outpainting. Both selected panoramas are then cropped using these identical parameter pairs to yield precisely aligned reference-target image pairs for the subsequent evaluation step.




(3) \textbf{Super-resolution and Filtering:} To mitigate the resolution degradation introduced during cropping, we deploy the OSEDiff  \cite{wu2024one} and RAM \cite{wang2022residual} models to  refine fine-grained details. Subsequently, we employ a Vision-Language Model (VLM) to evaluate each image pair across five criteria: geometric correlation, information balance, scene complexity, image clarity, and content coherence. The final score is strictly capped by the lowest sub-score. Detailed scoring guidelines and the full VLM prompt are available in the Supplementary Material.

(4) \textbf{Intermediate Sequence Generation:} To decompose large camera transformations into manageable steps, we generate $N$ intermediate frames for the target sequence. We achieve this by linearly interpolating the camera parameters between the input and output viewpoints to ensure temporal consistency.

\subsection{CamEditor-Bench Construction}
\label{sec:4.2}

To evaluate the generalization capability and robustness of camera-controlled editing, we introduce CamEditor-Bench. The primary objective of this benchmark is to provide a comprehensive evaluation suite spanning diverse semantic content, a broad spectrum of parameter variations, and strict data isolation.

\textbf{Scene and Domain Diversity.} To guarantee a rigorous evaluation of zero-shot generalization, CamEditor-Bench is curated to encompass a comprehensive semantic taxonomy while maintaining strict isolation from the training data. The evaluation panoramas are sampled from held-out splits and unseen environments across four domains: (1) Complex Indoor Environments, featuring intricate office layouts and building interiors; (2) Urban and Street Views, capturing dense cityscapes and first-person driving perspectives; (3) Natural Landscapes, encompassing unstructured outdoor scenery; and (4) Dynamic Human Activities, including extreme sports and crowds. This selection ensures the model is evaluated on novel geometric structures without scene-level data leakage.

\textbf{Input Form and Data Isolation.} Although panoramas provide the
ground-truth geometry needed to construct paired views, every benchmark input
is an ordinary perspective image cropped from a real-world or UE5 panorama.
CameraEditor therefore accepts a single perspective photograph at inference
time and does not require a panoramic user input. The training set (5,760
instances), CamEditor-Bench (462 held-out instances), and the inference-time
reference panorama pool are mutually disjoint. The reference pool contains
transformation exemplars rather than test-scene content and covers all
benchmark cases without scene-level overlap.

\textbf{Comprehensive Parameter Coverage.} We crop evaluation pairs from these base panoramas by sampling base and target viewpoint combinations that cover diverse structural perspective shifts and lens distortions, thereby reflecting unconstrained user inputs. Note that yaw is intentionally excluded from the evaluation; large yaw variations require extensive out-of-view content generation, shifting the task from controlled editing to outpainting.

\subsection{Metrics}
\label{sec:4.3}

To comprehensively evaluate the performance of CameraEditor on CamEditor-Bench, we establish a robust evaluation protocol divided into two primary dimensions: Image Quality \& Content Consistency, and Camera Geometric Accuracy.

\textbf{Image Quality \& Content Consistency.} We measure low-level structural and perceptual fidelity using SSIM \cite{wang2004image} and LPIPS \cite{zhang2018unreasonable}. To ensure the strict preservation of high-level semantic identity and 3D physical layout under extreme viewpoint shifts, we compute the cosine similarities of deep features extracted by CLIP (CLIP-I \cite{radford2021learning}) and DINO-v2 \cite{oquab2023dinov2}, respectively.
\textbf{Camera Geometric Accuracy.} Instead of estimating camera parameters separately, which typically requires non-linear optimization and introduces severe estimation bias, we propose an optimization-free evaluation approach based on the Perspective Field (PF). Specifically, we directly compute the dense representations: the \textit{upward vector field} and the \textit{latitude map}. We assess the geometric alignment through two distinct perspectives: 
\begin{itemize}
    \item \textbf{Alignment with GT Image (PF-I):} We compare the generated PF against the PF extracted from the ground-truth image to evaluate spatial structural preservation. 
    \item \textbf{Alignment with Target Conditions (PF-T):} We compare the generated PF directly against the user-provided target camera conditions in order to accurately measure instruction-following precision.
\end{itemize}
 For both perspectives, we quantify directional alignment via the cosine similarity of upward vectors ($S_{up}^I$, $S_{up}^T$) and absolute positioning via the Mean Squared Error of latitude values ($E_{lat}^I$, $E_{lat}^T$).


\section{Experiments}

\begin{figure*}[htbp]
    \centering
    \includegraphics[width=0.9\textwidth]{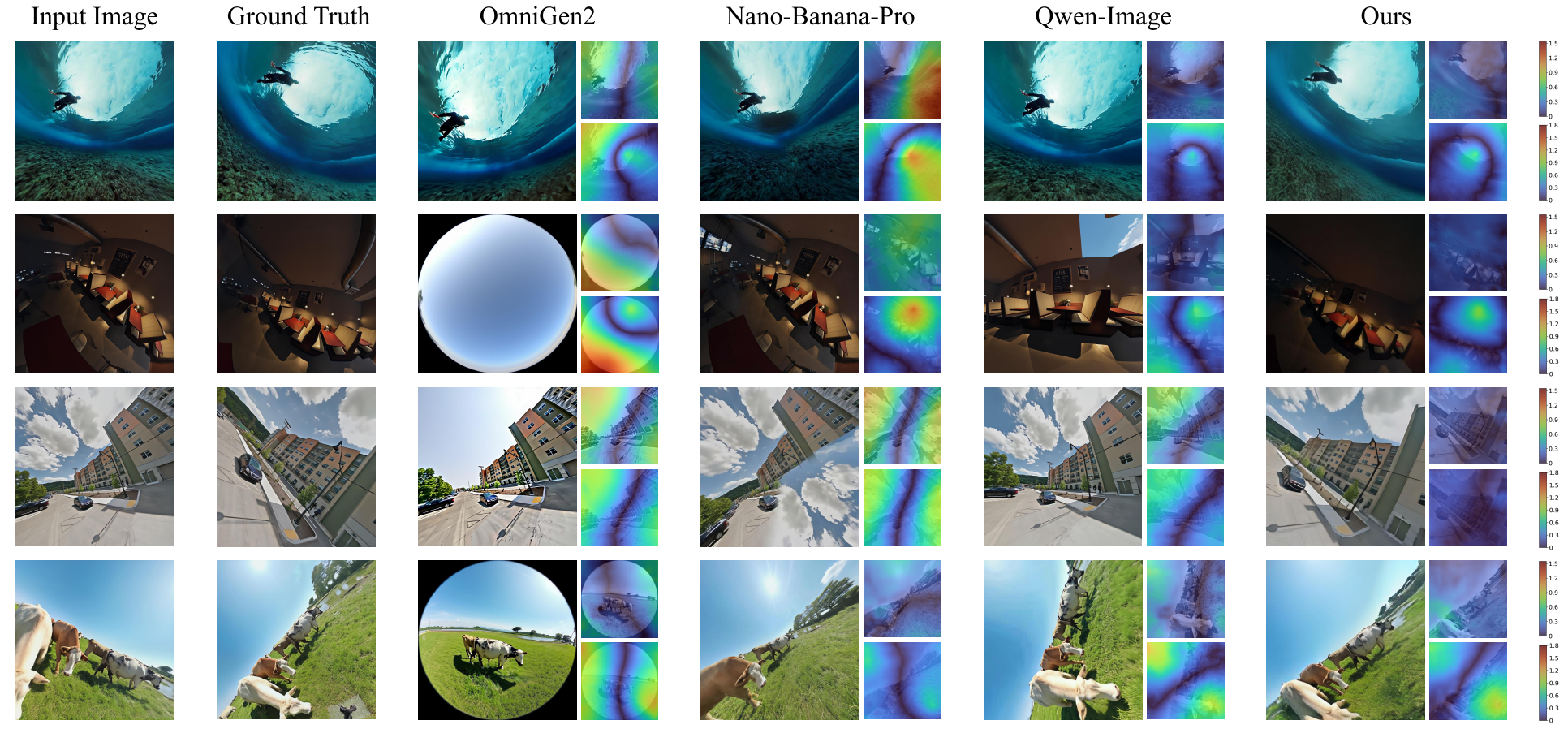}
    \caption{Qualitative results alongside geometric error maps. For each method, the right panels display PF-I (top) and PF-T (bottom) errors, where warmer colors indicate larger deviations.}
    \Description{Qualitative comparisons with PF-I and PF-T geometric error maps for each evaluated method.}
    \label{fig:main_results}
\end{figure*}

After detailing our experimental setup, we compare against state-of-the-art
baselines in Section~\ref{sec:rq1} to answer RQ1: Can temporal coherence
priors effectively break the trade-off between precise camera control and
content preservation? Section~\ref{sec:rq2} evaluates perception-module
robustness to answer RQ2: How does initial camera perception accuracy affect
final geometric alignment and visual quality? Finally,
Section~\ref{sec:ablation_sequence} answers RQ3: How does the number of
intermediate transition frames affect spatial coherence under camera
transformations?

\subsection{Experimental Setup}

CameraEditor uses the Wan2.1-T2V-14B video diffusion model
\cite{wan2025wan}. For the training data, we construct a specialized subset
comprising 5,760 instances. All input images and generated sequences are
processed at a spatial resolution of 512$\times$512. During fine-tuning, we
inject LoRA adapters with a rank of 32 into the attention blocks. The model is
optimized using AdamW (weight decay = 0.01) with a constant learning rate of
$1 \times 10^{-4}$ and a batch size of 4. During inference, based on our
ablation findings, we set the number of intermediate transition frames to
$N=8$. Consequently, the model generates a target sequence of exactly 9 frames
(8 intermediate frames and 1 final target frame).

To establish a comprehensive and rigorous baseline comparison, we evaluate CameraEditor against six open-source models (ICEdit \cite{zhang2025enabling}, Step1X-Edit \cite{liu2025step1x}, OmniGen2 \cite{wu2025omnigen2}, Flux.2 \cite{flux2dev}, Qwen-Image-Editing \cite{wu2025qwen}, and HunyuanImage-3.0-Instruct \cite{cao2025hunyuanimage}) and two closed-source proprietary APIs (GPT-Image-1.5 \cite{gptimage15} and Nano-Banana-Pro \cite{nanobananapro}). To accommodate the baseline models' reliance on natural-language instructions, we construct a standardized deterministic prompt-conversion pipeline. The exact mappings and baseline-selection rationale are provided in the Supplementary Material.

\textbf{Runtime.} On a single H200 GPU, GeoCalib requires approximately
68\,ms, dynamic reference routing approximately 7\,s, and 9-frame generation
approximately 46\,s, for a total latency of approximately 53\,s per input.
The generation stage dominates the runtime. Relative to random routing, the
7\,s routing stage improves DINO-v2 from 0.6301 to 0.8569 and reduces
$E_{lat}^{T}$ from 0.3712 to 0.1904
(Table~\ref{tab:perception_ablation}).

\begin{table*}[t]
\centering
\caption{\textbf{Quantitative comparison on CamEditor-Bench.} Evaluation of
image quality, content preservation, and camera geometric alignment. The two
Wan2.1 controls and CameraEditor use the same Wan2.1-T2V-14B backbone, training
data, and hyperparameters. $\uparrow$ indicates higher is better, and
$\downarrow$ indicates lower is better. Best results are in \textbf{bold}, and
second-best results are \underline{underlined}.}
\label{tab:main_results}

\renewcommand{\arraystretch}{0.85} 
\setlength{\tabcolsep}{4pt}

\resizebox{0.9\textwidth}{!}{
\small 
\begin{tabular}{lcccccccc} 
\toprule
\multirow{2}{*}{\textbf{Method}} & \multicolumn{4}{c}{\textbf{Content Preservation \& Quality}} & \multicolumn{4}{c}{\textbf{Camera Alignment}} \\
\cmidrule(lr){2-5} \cmidrule(lr){6-9}
& DINO-v2 ($\uparrow$) & CLIP-I ($\uparrow$) & SSIM ($\uparrow$) & LPIPS ($\downarrow$) & $S_{up}^I$ ($\uparrow$) & $E_{lat}^I$ ($\downarrow$) & $S_{up}^T$ ($\uparrow$) & $E_{lat}^T$ ($\downarrow$) \\
\midrule
ICEdit              & 0.7089 & 0.8009 & 0.3740 & 0.7161 & 0.8717 & 0.2032 & 0.5129 & 0.2928 \\
Step1X-Edit       & 0.7675 & 0.8502 & 0.4141 & 0.6838 & 0.8680 & 0.2061 & 0.5168 & 0.2972 \\
OmniGen2         & 0.6616 & 0.8041 & 0.3630 & 0.6774 & 0.8749 & 0.2029 & 0.5120 & 0.2937 \\
Flux.2    & 0.7604 & 0.8567 & 0.4173 & 0.6397 & 0.8806 & 0.1843 & 0.5295 & 0.2710 \\
  Qwen-Image-Editing         & 0.7749 & \underline{0.8723} & 0.4428 & 0.6345 & 0.8923 & \underline{0.1712} & 0.5354 & \underline{0.2549} \\
HunyuanImage-3.0-Instruct    & 0.7762 & 0.8417 & 0.3991 & 0.6551 & 0.8695 & 0.1910 & 0.5200 & 0.2816 \\
\midrule
GPT-Image-1.5      & 0.7703 & 0.8416 & 0.4469 & 0.6460 & 0.8524 & 0.2039 & 0.5153 & 0.2801 \\
  Nano-Banana-Pro     & \underline{0.7769} & 0.8642 & 0.4543 & 0.6440 & 0.8566 & 0.2136 & 0.5159 & 0.3020 \\
\midrule
\multicolumn{9}{l}{\textit{Same-backbone controls (Wan2.1-T2V-14B)}} \\
  Wan2.1, no reference & 0.6951 & 0.8235 & 0.4966 & 0.6380 & 0.8837 & 0.2091 & 0.5359 & 0.2802 \\
  Wan2.1 + Pl\"ucker & 0.7617 & 0.8652 & \underline{0.5139} & \underline{0.6340} & \underline{0.9186} & 0.1751 & \underline{0.5536} & 0.2575 \\
\rowcolor{gray!10} \textbf{CameraEditor (Ours)} & \textbf{0.8569} & \textbf{0.8970} & \textbf{0.5712} & \textbf{0.5454} & \textbf{0.9267} & \textbf{0.0917} & \textbf{0.6089} & \textbf{0.1904} \\
\bottomrule
\end{tabular}
}
\end{table*}

\subsection{Main Results (RQ1)}
\label{sec:rq1}

Evaluated under this rigorous framework, \textbf{CameraEditor achieves state-of-the-art performance across all metrics, effectively answering RQ1 by breaking the trade-off between precise geometric control and content preservation} (Table \ref{tab:main_results}). Compared to baselines, closed-source models preserve global aesthetics but struggle with strict 3D physical constraints, prioritizing stylistic priors over mathematical geometry. In contrast, instruction-driven open-source models reveal a clear performance hierarchy governed by model capacity. Small-scale models like OmniGen-v2 and ICEdit exhibit severe source identity loss; their limited representational bandwidth forces a destructive trade-off, rendering them unable to simultaneously decode semantics and compute large spatial transformations. Conversely, large-scale models (Flux.2, Qwen-Image, HunyuanImage-3.0) leverage richer pre-trained priors to mitigate structural collapse. However, their reliance on implicit textual guidance rather than explicit mathematical constraints fundamentally caps their absolute geometric precision. Among them, Qwen-Image achieves the most balanced empirical results (CLIP-I: 0.8723, $S_{up}^I$: 0.8923, $S_{up}^T$: 0.5354, and $E_{lat}^T$: 0.2549), yet remains below CameraEditor on both content preservation and target-camera alignment.

It is worth noting two specific metric divergences observed in our evaluation. First, most models maintain acceptable high-level semantic identity (DINO-v2 and CLIP-I) but show lower scores in low-level fidelity (SSIM and LPIPS). This is because massive perspective shifts inherently induce drastic, non-linear pixel displacements. Strictly aligned metrics like SSIM and LPIPS heavily penalize these natural structural deviations, even when the generated results are geometrically plausible and highly realistic. Second, relative geometric alignment with the ground truth ($S_{up}^I$, $E_{lat}^I$) consistently yields better scores than absolute alignment with the target camera parameters ($S_{up}^T$, $E_{lat}^T$). This numerical gap is driven by the systematic estimation deviation of current geometry-perception networks. Calculating relative alignment evaluates both generated and ground-truth images through the same network, effectively mitigating inherent domain biases, whereas absolute alignment strictly compares extracted features against pure mathematical conditions, exposing the perception network's intrinsic errors.

\subsection{Same-Backbone Controls and Human Evaluation}
\label{sec:fairness}

To isolate the effect of conditioning from backbone capacity, we train two
control variants with the same Wan2.1-T2V-14B backbone, training data, and
hyperparameters as CameraEditor. The \emph{no-reference} variant removes the
visual reference sequence and supplies the target camera through the same
deterministic prompt template used by the instruction-driven baselines. The
\emph{Pl\"ucker} variant replaces the visual reference with explicit
Pl\"ucker camera embeddings derived from GeoCalib. CoF is retained in both
variants. The controls are integrated directly into the main comparison
(Table~\ref{tab:main_results}). Moving from no reference to Pl\"ucker
conditioning improves DINO-v2 from 0.6951 to 0.7617 and
$S_{up}^I$ from 0.8837 to 0.9186; the full visual-reference formulation
further reaches 0.8569 and 0.9267, respectively. These controlled results
demonstrate that the gain is not attributable to backbone capacity alone.

We further conduct 300 blind pairwise human comparisons against Step1X-Edit,
Qwen-Image-Editing, and Nano-Banana-Pro (100 comparisons per baseline). Each
comparison presents the source image, ground truth, target-camera
visualization, and two anonymous outputs. Evaluators select the better edit
based jointly on perceived camera correctness, content preservation, and
artifact severity. As shown in Table~\ref{tab:human_study}, CameraEditor is
preferred in 82.0\%, 63.5\%, and 69.5\% of the comparisons, respectively,
when each tie contributes 0.5 to both methods. The human ranking is
consistent with the automatic camera-alignment and content-preservation
metrics.

\begin{table}[t]
\centering
\caption{\textbf{Blind pairwise human evaluation.} Preference counts a tie as
0.5 for each method. Each row contains 100 comparisons.}
\label{tab:human_study}
\setlength{\tabcolsep}{3.5pt}
\begin{tabular}{lrrrr}
\toprule
\textbf{Baseline} & \textbf{Base} & \textbf{Tie} & \textbf{Ours} &
\textbf{Pref.} \\
\midrule
Step1X-Edit & 11 & 14 & 75 & 82.0\% \\
Qwen-Image-Editing & 25 & 23 & 52 & 63.5\% \\
Nano-Banana-Pro & 19 & 23 & 58 & 69.5\% \\
\bottomrule
\end{tabular}
\end{table}

\begin{table*}[t]
\centering
\caption{\textbf{Ablation study on Perception and Routing Module.} Comparison of different reference routing strategies. Relying on our dedicated perception module (GeoCalib) provides the most accurate geometric anchor and best visual quality.}
\label{tab:perception_ablation}
\renewcommand{\arraystretch}{0.85} 
\setlength{\tabcolsep}{4pt}
\resizebox{0.8\textwidth}{!}{
\small 
\begin{tabular}{lcccccccc} 
\toprule
\multirow{2}{*}{\textbf{Routing Strategy}} & \multicolumn{4}{c}{\textbf{Content Preservation \& Quality}} & \multicolumn{4}{c}{\textbf{Camera Alignment}} \\
\cmidrule(lr){2-5} \cmidrule(lr){6-9}
& DINO-v2 ($\uparrow$) & CLIP-I ($\uparrow$) & SSIM ($\uparrow$) & LPIPS ($\downarrow$) & $S_{up}^I$ ($\uparrow$) & $E_{lat}^I$ ($\downarrow$) & $S_{up}^T$ ($\uparrow$) & $E_{lat}^T$ ($\downarrow$) \\
\midrule
Random          & 0.6301 & 0.8083 & 0.5231 & 0.6281 & 0.8419 & 0.2654 & 0.4961 & 0.3712 \\
Puffin base     & 0.7413 & 0.8486 & 0.5360 & 0.6009 & 0.8936 & 0.1548 & \underline{0.5752} & \underline{0.2329} \\
Puffin thinking & \underline{0.7478} & \underline{0.8508} & \underline{0.5454} & \underline{0.5993} & \underline{0.9049} & \underline{0.1523} & 0.5652 & 0.2439 \\
\midrule
\rowcolor{gray!10} \textbf{GeoCalib} \cite{veicht2024geocalib} & \textbf{0.8569} & \textbf{0.8970} & \textbf{0.5712} & \textbf{0.5454} & \textbf{0.9267} & \textbf{0.0917} & \textbf{0.6089} & \textbf{0.1904} \\
\bottomrule
\end{tabular}
}
\end{table*}

\begin{table*}[t]
\centering
\caption{\textbf{Ablation study on the Number of Intermediate Frames (N).} The impact of intermediate transition frames on spatial coherence, content preservation, and camera alignment.}
\label{tab:sequence_length}
\renewcommand{\arraystretch}{0.85} 
\setlength{\tabcolsep}{4pt}
\resizebox{0.8\textwidth}{!}{
\small 
\begin{tabular}{lcccccccc} 
\toprule
\multirow{2}{*}{\textbf{Sequence Length}} & \multicolumn{4}{c}{\textbf{Content Preservation \& Quality}} & \multicolumn{4}{c}{\textbf{Camera Alignment}} \\
\cmidrule(lr){2-5} \cmidrule(lr){6-9}
& DINO-v2 ($\uparrow$) & CLIP-I ($\uparrow$) & SSIM ($\uparrow$) & LPIPS ($\downarrow$) & $S_{up}^I$ ($\uparrow$) & $E_{lat}^I$ ($\downarrow$) & $S_{up}^T$ ($\uparrow$) & $E_{lat}^T$ ($\downarrow$) \\
\midrule
N=0 & \underline{0.8435} & 0.8686 & 0.5181 & 0.5891 & 0.9010 & 0.1343 & \textbf{0.6242} & \textbf{0.1669} \\
N=4 & 0.8128 & 0.8766 & 0.5559 & \underline{0.5645} & 0.9223 & 0.1184 & \underline{0.6231} & 0.1974 \\
\rowcolor{gray!10} \textbf{N=8} & \textbf{0.8569} & \textbf{0.8970} & \textbf{0.5712} & \textbf{0.5454} & \underline{0.9267} & \textbf{0.0917} & 0.6089 & \underline{0.1904} \\
N=12 & 0.8179 & \underline{0.8781} & \underline{0.5604} & 0.5925 & \textbf{0.9286} & \underline{0.1110} & 0.5824 & 0.2476 \\
\bottomrule
\end{tabular}
}
\end{table*}

\subsection{Ablation Study on Perception Module (RQ2)}
\label{sec:rq2}

To answer RQ2, we evaluate the impact of the initial camera perception module
on final editing quality. Reference sequences are analytically cropped under
known camera parameters; routing errors arise from estimating the input
configuration and matching it to the candidate pool. We therefore evaluate
routing quality through its effect on the generated final frame. More accurate
initial perception yields a reference sequence that more closely corresponds
to the target trajectory, improving viewpoint consistency across frames.
Detailed parameter-estimation errors and visual comparisons are provided in
the Supplementary Material.

As reported in Table~\ref{tab:perception_ablation}, random routing yields the
poorest performance. Without an accurate geometric anchor, the reference
sequence fails to correspond with the target, destroying frame-to-frame
viewpoint consistency and causing spatial collapse across the temporal
sequence ($E_{lat}^I$: 0.2654, DINO-v2: 0.6301).

We then evaluate Puffin \cite{liao2025thinking} (both base and thinking variants), which represents VLM-based camera understanding. While Puffin provides a rough sequence correspondence, it struggles to accurately regress rigorous mathematical parameters, particularly non-linear optical distortions. Consequently, the reference and target frames suffer from perceptual deviation. This misalignment inherently caps the absolute geometric accuracy (e.g., Puffin thinking achieves an $S_{up}^T$ of only 0.5652) and degrades source content preservation (DINO-v2: 0.7478).

In contrast, our pipeline leverages GeoCalib to estimate the camera
configuration and retrieve a more closely aligned reference sequence. As
shown in Table~\ref{tab:perception_ablation}, this sequence-level alignment
reduces relative geometric errors and yields the highest absolute alignment
and content-preservation scores among the evaluated routing strategies.

\subsection{Ablation Study on Intermediate Frame Counts (RQ3)}

\label{sec:ablation_sequence}

To address RQ3, we evaluate the impact of intermediate transition frame counts ($N \in \{0, 4, 8, 12\}$) on editing performance (Table \ref{tab:sequence_length}). All variants use the same Wan2.1-T2V-14B backbone, training data, and conditioning pipeline; consequently, $N=0$ is the requested single-step same-backbone control. The resulting total sequence lengths ($N+1 \in \{1, 5, 9, 13\}$) are strictly chosen to satisfy the $4n+1$ temporal constraint of the Wan2.1 3D-VAE backbone. This architectural alignment obviates the need for temporal padding during encoding, inherently preventing latent misalignment and subsequent interpolation artifacts.

\textbf{Results reveal a strict trade-off between absolute geometric adherence and structural integrity}. Under a constrained temporal window ($N=0$), tight cross-frame coupling forces aggressive latent warping. This yields the highest absolute geometric alignment ($S_{up}^T$: 0.6242, $E_{lat}^T$: 0.1669) but induces severe structural tearing due to massive perspective shifts in limited steps, resulting in the poorest low-level fidelity (SSIM: 0.5181, LPIPS: 0.5891).

Conversely, increasing $N$ mitigates this tearing by decomposing spatial transformations into a temporal buffer. However, excessive extension (e.g., $N=12$) introduces \textit{temporal drift}. While smoother transitions maintain high relative structural consistency ($S_{up}^I$), the prolonged iterative denoising process accumulates latent noise. This progressive dilution weakens the mathematical anchors at the endpoints, reducing absolute alignment ($S_{up}^T$: 0.5824, $E_{lat}^T$: 0.2476) and degrading content preservation (DINO-v2: 0.8179).

Consequently, $N=8$ provides the optimal balance. It offers temporal buffering to prevent structural tearing---achieving peak visual quality (SSIM: 0.5712, LPIPS: 0.5454) and content fidelity (DINO-v2: 0.8569)---while averting temporal drift to maintain precise relative geometric alignment ($E_{lat}^I$: 0.0917).


\section{Conclusions and Limitations}

We present CameraEditor for camera-parameterized image editing. Our framework formulates the problem as a temporal sequence prediction task, effectively leveraging the temporal coherence priors of video diffusion models. By integrating explicit geometric perception with dynamic panorama cropping, our approach establishes robust geometric anchors for complex spatial manipulations. Additionally, the strategic insertion of intermediate transition frames effectively decomposes large-scale perspective shifts, mitigating potential structural distortions and buffering initial observation errors. Extensive evaluations on the newly introduced CamEditor-Bench demonstrate that CameraEditor achieves state-of-the-art performance, striking a highly effective balance between precise spatial alignment and source content preservation.

Limitations and Future Work. Despite its effectiveness, our approach has two primary limitations. First, the inherent spatial compression within video 3D-VAEs tends to over-smooth high-frequency details, leading to the degradation of fine-grained textures during the decoding process. Second, the current system operates as a cascaded pipeline---sequentially executing parameter extraction, reference cropping, and video denoising. This decoupled design lacks end-to-end optimization and increases inference latency. To address these issues, future work will explore hybrid latent upscaling for robust texture recovery and a unified, end-to-end architecture for efficient camera-controlled editing.

\begin{acks}
\begingroup
\emergencystretch=2em
This work was supported by the New Generation Artificial
Intelligence National Science and Technology Major Project
(No.~2025ZD0123001), the National Natural Science Foundation of China
(Nos.~62272374 and 62192781), the Natural Science Foundation of
Shaanxi Province (No.~2024JC-JCQN-62), the State Key Laboratory of
Communication Content Cognition under Grant No.~A202502, and the Key
Research and Development Project in Shaanxi Province
(No.~2023GXLH-024).

This work was also supported by the A*STAR Career Development Fund
(Project No.~C243512010).
\par
\endgroup
\end{acks}

\bibliographystyle{ACM-Reference-Format}
\bibliography{references}

@String{Computer = "{IEEE} Computer" }

@String{Springer = "Springer-Verlag" }

@article{zhang2025unified,
  title   = {Unified multimodal understanding and generation models: Advances, challenges, and opportunities},
  author  = {Zhang, Xinjie and Guo, Jintao and Zhao, Shanshan and Fu, Minghao and Duan, Lunhao and Hu, Jiakui and Chng, Yong Xien and Wang, Guo-Hua and Chen, Qing-Guo and Xu, Zhao and others},
  journal = {arXiv preprint arXiv:2505.02567},
  year    = {2025}
}

@inproceedings{brooks2023instructpix2pix,
  title     = {Instructpix2pix: Learning to follow image editing instructions},
  author    = {Brooks, Tim and Holynski, Aleksander and Efros, Alexei A},
  booktitle = {Proceedings of the IEEE/CVF conference on computer vision and pattern recognition},
  pages     = {18392--18402},
  year      = {2023}
}

@article{betker2023improving,
  title   = {Improving image generation with better captions},
  author  = {Betker, James and Goh, Gabriel and Jing, Li and Brooks, Tim and Wang, Jianfeng and Li, Linjie and Ouyang, Long and Zhuang, Juntang and Lee, Joyce and Guo, Yufei and others},
  journal = {Computer Science. https://cdn. openai. com/papers/dall-e-3. pdf},
  volume  = {2},
  number  = {3},
  pages   = {8},
  year    = {2023}
}

@inproceedings{rombach2022high,
  title     = {High-resolution image synthesis with latent diffusion models},
  author    = {Rombach, Robin and Blattmann, Andreas and Lorenz, Dominik and Esser, Patrick and Ommer, Bj{\"o}rn},
  booktitle = {Proceedings of the IEEE/CVF conference on computer vision and pattern recognition},
  pages     = {10684--10695},
  year      = {2022}
}

@article{huang2024context,
  title   = {In-context lora for diffusion transformers},
  author  = {Huang, Lianghua and Wang, Wei and Wu, Zhi-Fan and Shi, Yupeng and Dou, Huanzhang and Liang, Chen and Feng, Yutong and Liu, Yu and Zhou, Jingren},
  journal = {arXiv preprint arXiv:2410.23775},
  year    = {2024}
}

@article{xu2023cyclenet,
  title   = {Cyclenet: Rethinking cycle consistency in text-guided diffusion for image manipulation},
  author  = {Xu, Sihan and Ma, Ziqiao and Huang, Yidong and Lee, Honglak and Chai, Joyce},
  journal = {Advances in Neural Information Processing Systems},
  volume  = {36},
  pages   = {10359--10384},
  year    = {2023}
}

@article{zhang2025context,
  title   = {In-context edit: Enabling instructional image editing with in-context generation in large scale diffusion transformer},
  author  = {Zhang, Zechuan and Xie, Ji and Lu, Yu and Yang, Zongxin and Yang, Yi},
  journal = {arXiv preprint arXiv:2504.20690},
  year    = {2025}
}

@inproceedings{ruiz2023dreambooth,
  title     = {Dreambooth: Fine tuning text-to-image diffusion models for subject-driven generation},
  author    = {Ruiz, Nataniel and Li, Yuanzhen and Jampani, Varun and Pritch, Yael and Rubinstein, Michael and Aberman, Kfir},
  booktitle = {Proceedings of the IEEE/CVF conference on computer vision and pattern recognition},
  pages     = {22500--22510},
  year      = {2023}
}

@article{huang2025diffusion,
  title     = {Diffusion model-based image editing: A survey},
  author    = {Huang, Yi and Huang, Jiancheng and Liu, Yifan and Yan, Mingfu and Lv, Jiaxi and Liu, Jianzhuang and Xiong, Wei and Zhang, He and Cao, Liangliang and Chen, Shifeng},
  journal   = {IEEE Transactions on Pattern Analysis and Machine Intelligence},
  year      = {2025},
  publisher = {IEEE}
}

@article{rout2024semantic,
  title   = {Semantic image inversion and editing using rectified stochastic differential equations},
  author  = {Rout, Litu and Chen, Yujia and Ruiz, Nataniel and Caramanis, Constantine and Shakkottai, Sanjay and Chu, Wen-Sheng},
  journal = {arXiv preprint arXiv:2410.10792},
  year    = {2024}
}

@inproceedings{yu2025anyedit,
  title     = {Anyedit: Mastering unified high-quality image editing for any idea},
  author    = {Yu, Qifan and Chow, Wei and Yue, Zhongqi and Pan, Kaihang and Wu, Yang and Wan, Xiaoyang and Li, Juncheng and Tang, Siliang and Zhang, Hanwang and Zhuang, Yueting},
  booktitle = {Proceedings of the Computer Vision and Pattern Recognition Conference},
  pages     = {26125--26135},
  year      = {2025}
}

@article{huang2025pfb,
  title     = {Pfb-diff: Progressive feature blending diffusion for text-driven image editing},
  author    = {Huang, Wenjing and Tu, Shikui and Xu, Lei},
  journal   = {Neural Networks},
  volume    = {181},
  pages     = {106777},
  year      = {2025},
  publisher = {Elsevier}
}

@inproceedings{han2025stylebooth,
  title     = {Stylebooth: Image style editing with multimodal instruction},
  author    = {Han, Zhen and Mao, Chaojie and Jiang, Zeyinzi and Pan, Yulin and Zhang, Jingfeng},
  booktitle = {Proceedings of the IEEE/CVF International Conference on Computer Vision},
  pages     = {1947--1957},
  year      = {2025}
}

@inproceedings{lei2025stylestudio,
  title={Stylestudio: Text-driven style transfer with selective control of style elements},
  author={Lei, Mingkun and Song, Xue and Zhu, Beier and Wang, Hao and Zhang, Chi},
  booktitle={Proceedings of the Computer Vision and Pattern Recognition Conference},
  pages={23443--23452},
  year={2025}
}

@article{yan2025eedit,
  title   = {Eedit: Rethinking the spatial and temporal redundancy for efficient image editing},
  author  = {Yan, Zexuan and Ma, Yue and Zou, Chang and Chen, Wenteng and Chen, Qifeng and Zhang, Linfeng},
  journal = {arXiv preprint arXiv:2503.10270},
  year    = {2025}
}

@article{wu2025qwen,
  title   = {Qwen-image technical report},
  author  = {Wu, Chenfei and Li, Jiahao and Zhou, Jingren and Lin, Junyang and Gao, Kaiyuan and Yan, Kun and Yin, Sheng-ming and Bai, Shuai and Xu, Xiao and Chen, Yilei and others},
  journal = {arXiv preprint arXiv:2508.02324},
  year    = {2025}
}

@inproceedings{jin2023perspective,
  title     = {Perspective fields for single image camera calibration},
  author    = {Jin, Linyi and Zhang, Jianming and Hold-Geoffroy, Yannick and Wang, Oliver and Blackburn-Matzen, Kevin and Sticha, Matthew and Fouhey, David F},
  booktitle = {Proceedings of the IEEE/CVF Conference on Computer Vision and Pattern Recognition},
  pages     = {17307--17316},
  year      = {2023}
}

@article{kong2024hunyuanvideo,
  title   = {Hunyuanvideo: A systematic framework for large video generative models},
  author  = {Kong, Weijie and Tian, Qi and Zhang, Zijian and Min, Rox and Dai, Zuozhuo and Zhou, Jin and Xiong, Jiangfeng and Li, Xin and Wu, Bo and Zhang, Jianwei and others},
  journal = {arXiv preprint arXiv:2412.03603},
  year    = {2024}
}

@article{ma2024latte,
  title   = {Latte: Latent diffusion transformer for video generation},
  author  = {Ma, Xin and Wang, Yaohui and Jia, Gengyun and Chen, Xinyuan and Liu, Ziwei and Li, Yuan-Fang and Chen, Cunjian and Qiao, Yu},
  journal = {arXiv preprint arXiv:2401.03048},
  year    = {2024}
}

@article{lu2023vdt,
  title   = {Vdt: General-purpose video diffusion transformers via mask modeling},
  author  = {Lu, Haoyu and Yang, Guoxing and Fei, Nanyi and Huo, Yuqi and Lu, Zhiwu and Luo, Ping and Ding, Mingyu},
  journal = {arXiv preprint arXiv:2305.13311},
  year    = {2023}
}

@article{el2022unreal,
  title     = {Unreal Engine 5 and immersive surgical training: translating advances in gaming technology into extended-reality surgical simulation training programmes},
  author    = {El-Wajeh, Yasin AM and Hatton, Paul V and Lee, Nicholas J},
  journal   = {British Journal of Surgery},
  volume    = {109},
  number    = {5},
  pages     = {470--471},
  year      = {2022},
  publisher = {Oxford University Press}
}

@article{pollefeys1999self,
  title     = {Self-calibration and metric reconstruction inspite of varying and unknown intrinsic camera parameters},
  author    = {Pollefeys, Marc and Koch, Reinhard and Gool, Luc Van},
  journal   = {International journal of computer vision},
  volume    = {32},
  number    = {1},
  pages     = {7--25},
  year      = {1999},
  publisher = {Springer}
}

@book{hartley2003multiple,
  title     = {Multiple view geometry in computer vision},
  author    = {Hartley, Richard and Zisserman, Andrew},
  year      = {2003},
  publisher = {Cambridge university press}
}

@inproceedings{sinha2023sparsepose,
  title     = {Sparsepose: Sparse-view camera pose regression and refinement},
  author    = {Sinha, Samarth and Zhang, Jason Y and Tagliasacchi, Andrea and Gilitschenski, Igor and Lindell, David B},
  booktitle = {Proceedings of the IEEE/CVF Conference on Computer Vision and Pattern Recognition},
  pages     = {21349--21359},
  year      = {2023}
}

@inproceedings{hold2018perceptual,
  title     = {A perceptual measure for deep single image camera calibration},
  author    = {Hold-Geoffroy, Yannick and Sunkavalli, Kalyan and Eisenmann, Jonathan and Fisher, Matthew and Gambaretto, Emiliano and Hadap, Sunil and Lalonde, Jean-Fran{\c{c}}ois},
  booktitle = {Proceedings of the IEEE Conference on Computer Vision and Pattern Recognition},
  pages     = {2354--2363},
  year      = {2018}
}

@inproceedings{workman2015deepfocal,
  title        = {Deepfocal: A method for direct focal length estimation},
  author       = {Workman, Scott and Greenwell, Connor and Zhai, Menghua and Baltenberger, Ryan and Jacobs, Nathan},
  booktitle    = {2015 IEEE International Conference on Image Processing (ICIP)},
  pages        = {1369--1373},
  year         = {2015},
  organization = {IEEE}
}

@article{liao2020model,
  title     = {Model-free distortion rectification framework bridged by distortion distribution map},
  author    = {Liao, Kang and Lin, Chunyu and Zhao, Yao and Xu, Mai},
  journal   = {IEEE Transactions on Image Processing},
  volume    = {29},
  pages     = {3707--3718},
  year      = {2020},
  publisher = {IEEE}
}

@inproceedings{zhuang2021fusing,
  title     = {Fusing the old with the new: Learning relative camera pose with geometry-guided uncertainty},
  author    = {Zhuang, Bingbing and Chandraker, Manmohan},
  booktitle = {Proceedings of the IEEE/CVF Conference on Computer Vision and Pattern Recognition},
  pages     = {32--42},
  year      = {2021}
}

@article{zheng2024cami2v,
  title   = {Cami2v: Camera-controlled image-to-video diffusion model},
  author  = {Zheng, Guangcong and Li, Teng and Jiang, Rui and Lu, Yehao and Wu, Tao and Li, Xi},
  journal = {arXiv preprint arXiv:2410.15957},
  year    = {2024}
}

@article{he2024cameractrl,
  title   = {Cameractrl: Enabling camera control for text-to-video generation},
  author  = {He, Hao and Xu, Yinghao and Guo, Yuwei and Wetzstein, Gordon and Dai, Bo and Li, Hongsheng and Yang, Ceyuan},
  journal = {arXiv preprint arXiv:2404.02101},
  year    = {2024}
}

@inproceedings{zhang2020fixation,
  title        = {A fixation-based 360 benchmark dataset for salient object detection},
  author       = {Zhang, Yi and Zhang, Lu and Hamidouche, Wassim and Deforges, Olivier},
  booktitle    = {2020 IEEE International Conference on Image Processing (ICIP)},
  pages        = {3458--3462},
  year         = {2020},
  organization = {IEEE}
}

@article{orhan2022semantic,
  title     = {Semantic segmentation of outdoor panoramic images},
  author    = {Orhan, Semih and Bastanlar, Yalin},
  journal   = {Signal, Image and Video Processing},
  volume    = {16},
  number    = {3},
  pages     = {643--650},
  year      = {2022},
  publisher = {Springer}
}

@article{li2019distortion,
  title     = {Distortion-Adaptive Salient Object Detection in 360$^\circ$ Omnidirectional Images},
  author    = {Li, Jia and Su, Jinming and Xia, Changqun and Tian, Yonghong},
  journal   = {IEEE Journal of Selected Topics in Signal Processing},
  volume    = {14},
  number    = {1},
  pages     = {38--48},
  year      = {2019},
  publisher = {IEEE}
}

@article{ghazanfari2025chain,
  title={Chain-of-frames: Advancing video understanding in multimodal llms via frame-aware reasoning},
  author={Ghazanfari, Sara and Croce, Francesco and Flammarion, Nicolas and Krishnamurthy, Prashanth and Khorrami, Farshad and Garg, Siddharth},
  journal={arXiv preprint arXiv:2506.00318},
  year={2025}
}

@inproceedings{tomczak2018vae,
  title        = {VAE with a VampPrior},
  author       = {Tomczak, Jakub and Welling, Max},
  booktitle    = {International conference on artificial intelligence and statistics},
  pages        = {1214--1223},
  year         = {2018},
  organization = {PMLR}
}

@article{lipman2022flow,
  title   = {Flow matching for generative modeling},
  author  = {Lipman, Yaron and Chen, Ricky TQ and Ben-Hamu, Heli and Nickel, Maximilian and Le, Matt},
  journal = {arXiv preprint arXiv:2210.02747},
  year    = {2022}
}

@article{wu2024one,
  title   = {One-step effective diffusion network for real-world image super-resolution},
  author  = {Wu, Rongyuan and Sun, Lingchen and Ma, Zhiyuan and Zhang, Lei},
  journal = {Advances in Neural Information Processing Systems},
  volume  = {37},
  pages   = {92529--92553},
  year    = {2024}
}

@article{wang2022residual,
  title     = {Residual deep attention mechanism and adaptive reconstruction network for single image super-resolution},
  author    = {Wang, Hongjuan and Wei, Mingrun and Cheng, Ru and Yu, Yue and Zhang, Xingli},
  journal   = {Applied Intelligence},
  volume    = {52},
  number    = {5},
  pages     = {5197--5211},
  year      = {2022},
  publisher = {Springer}
}

@inproceedings{bernal2025precisecam,
  title     = {PreciseCam: Precise Camera Control for Text-to-Image Generation},
  author    = {Bernal-Berdun, Edurne and Serrano, Ana and Masia, Belen and Gadelha, Matheus and Hold-Geoffroy, Yannick and Sun, Xin and Gutierrez, Diego},
  booktitle = {Proceedings of the Computer Vision and Pattern Recognition Conference},
  pages     = {2724--2733},
  year      = {2025}
}

@article{wang2004image,
  title     = {Image quality assessment: from error visibility to structural similarity},
  author    = {Wang, Zhou and Bovik, Alan C and Sheikh, Hamid R and Simoncelli, Eero P},
  journal   = {IEEE transactions on image processing},
  volume    = {13},
  number    = {4},
  pages     = {600--612},
  year      = {2004},
  publisher = {IEEE}
}

@inproceedings{zhang2018unreasonable,
  title     = {The unreasonable effectiveness of deep features as a perceptual metric},
  author    = {Zhang, Richard and Isola, Phillip and Efros, Alexei A and Shechtman, Eli and Wang, Oliver},
  booktitle = {Proceedings of the IEEE conference on computer vision and pattern recognition},
  pages     = {586--595},
  year      = {2018}
}

@inproceedings{radford2021learning,
  title        = {Learning transferable visual models from natural language supervision},
  author       = {Radford, Alec and Kim, Jong Wook and Hallacy, Chris and Ramesh, Aditya and Goh, Gabriel and Agarwal, Sandhini and Sastry, Girish and Askell, Amanda and Mishkin, Pamela and Clark, Jack and others},
  booktitle    = {International conference on machine learning},
  pages        = {8748--8763},
  year         = {2021},
  organization = {PmLR}
}

@article{oquab2023dinov2,
  title   = {Dinov2: Learning robust visual features without supervision},
  author  = {Oquab, Maxime and Darcet, Timoth{\'e}e and Moutakanni, Th{\'e}o and Vo, Huy and Szafraniec, Marc and Khalidov, Vasil and Fernandez, Pierre and Haziza, Daniel and Massa, Francisco and El-Nouby, Alaaeldin and others},
  journal = {arXiv preprint arXiv:2304.07193},
  year    = {2023}
}

@article{liao2025thinking,
  title   = {Thinking with Camera: A Unified Multimodal Model for Camera-Centric Understanding and Generation},
  author  = {Liao, Kang and Wu, Size and Wu, Zhonghua and Jin, Linyi and Wang, Chao and Wang, Yikai and Wang, Fei and Li, Wei and Loy, Chen Change},
  journal = {arXiv preprint arXiv:2510.08673},
  year    = {2025}
}

@article{hurst2024gpt,
  title   = {Gpt-4o system card},
  author  = {Hurst, Aaron and Lerer, Adam and Goucher, Adam P and Perelman, Adam and Ramesh, Aditya and Clark, Aidan and Ostrow, AJ and Welihinda, Akila and Hayes, Alan and Radford, Alec and others},
  journal = {arXiv preprint arXiv:2410.21276},
  year    = {2024}
}

@article{imran2024google,
  title     = {Google Gemini as a next generation AI educational tool: a review of emerging educational technology},
  author    = {Imran, Muhammad and Almusharraf, Norah},
  journal   = {Smart Learning Environments},
  volume    = {11},
  number    = {1},
  pages     = {22},
  year      = {2024},
  publisher = {Springer}
}

@article{tsidylo2023artificial,
  title     = {Artificial intelligence as a methodological innovation in the training of future designers: Midjourney tools},
  author    = {Tsidylo, Ivan M and Sena, Chele Esteve},
  journal   = {Information Technologies and Learning Tools},
  volume    = {97},
  number    = {5},
  pages     = {203},
  year      = {2023},
  publisher = {Institute for Digitalisation of Education of the National Academy of~…}
}

@inproceedings{esser2023structure,
  title     = {Structure and content-guided video synthesis with diffusion models},
  author    = {Esser, Patrick and Chiu, Johnathan and Atighehchian, Parmida and Granskog, Jonathan and Germanidis, Anastasis},
  booktitle = {Proceedings of the IEEE/CVF international conference on computer vision},
  pages     = {7346--7356},
  year      = {2023}
}

@inproceedings{zhang2023adding,
  title     = {Adding conditional control to text-to-image diffusion models},
  author    = {Zhang, Lvmin and Rao, Anyi and Agrawala, Maneesh},
  booktitle = {Proceedings of the IEEE/CVF international conference on computer vision},
  pages     = {3836--3847},
  year      = {2023}
}

@inproceedings{mou2024t2i,
  title     = {T2i-adapter: Learning adapters to dig out more controllable ability for text-to-image diffusion models},
  author    = {Mou, Chong and Wang, Xintao and Xie, Liangbin and Wu, Yanze and Zhang, Jian and Qi, Zhongang and Shan, Ying},
  booktitle = {Proceedings of the AAAI conference on artificial intelligence},
  volume    = {38},
  number    = {5},
  pages     = {4296--4304},
  year      = {2024}
}

@inproceedings{li2023gligen,
  title     = {Gligen: Open-set grounded text-to-image generation},
  author    = {Li, Yuheng and Liu, Haotian and Wu, Qingyang and Mu, Fangzhou and Yang, Jianwei and Gao, Jianfeng and Li, Chunyuan and Lee, Yong Jae},
  booktitle = {Proceedings of the IEEE/CVF conference on computer vision and pattern recognition},
  pages     = {22511--22521},
  year      = {2023}
}

@inproceedings{qin2025camedit,
  title     = {CamEdit: Continuous Camera Parameter Control for Photorealistic Image Editing},
  author    = {Qin, Xinran and Wang, Zhixin and Li, Fan and Chen, Haoyu and Pei, RenJing and Li, WenBo and Cao, XiaoChun},
  booktitle = {The Thirty-ninth Annual Conference on Neural Information Processing Systems},
  year      = {2025}
}

@inproceedings{schonberger2016structure,
  title     = {Structure-from-motion revisited},
  author    = {Schonberger, Johannes L and Frahm, Jan-Michael},
  booktitle = {Proceedings of the IEEE conference on computer vision and pattern recognition},
  pages     = {4104--4113},
  year      = {2016}
}

@inproceedings{kendall2015posenet,
  title     = {Posenet: A convolutional network for real-time 6-dof camera relocalization},
  author    = {Kendall, Alex and Grimes, Matthew and Cipolla, Roberto},
  booktitle = {Proceedings of the IEEE international conference on computer vision},
  pages     = {2938--2946},
  year      = {2015}
}

@article{mildenhall2021nerf,
  title     = {Nerf: Representing scenes as neural radiance fields for view synthesis},
  author    = {Mildenhall, Ben and Srinivasan, Pratul P and Tancik, Matthew and Barron, Jonathan T and Ramamoorthi, Ravi and Ng, Ren},
  journal   = {Communications of the ACM},
  volume    = {65},
  number    = {1},
  pages     = {99--106},
  year      = {2021},
  publisher = {ACM New York, NY, USA}
}

@inproceedings{liu2023zero,
  title     = {Zero-1-to-3: Zero-shot one image to 3d object},
  author    = {Liu, Ruoshi and Wu, Rundi and Van Hoorick, Basile and Tokmakov, Pavel and Zakharov, Sergey and Vondrick, Carl},
  booktitle = {Proceedings of the IEEE/CVF international conference on computer vision},
  pages     = {9298--9309},
  year      = {2023}
}

@article{shi2023zero123++,
  title   = {Zero123++: a single image to consistent multi-view diffusion base model},
  author  = {Shi, Ruoxi and Chen, Hansheng and Zhang, Zhuoyang and Liu, Minghua and Xu, Chao and Wei, Xinyue and Chen, Linghao and Zeng, Chong and Su, Hao},
  journal = {arXiv preprint arXiv:2310.15110},
  year    = {2023}
}

@inproceedings{wang2024motionctrl,
  title     = {Motionctrl: A unified and flexible motion controller for video generation},
  author    = {Wang, Zhouxia and Yuan, Ziyang and Wang, Xintao and Li, Yaowei and Chen, Tianshui and Xia, Menghan and Luo, Ping and Shan, Ying},
  booktitle = {ACM SIGGRAPH 2024 Conference Papers},
  pages     = {1--11},
  year      = {2024}
}

@inproceedings{veicht2024geocalib,
  title        = {Geocalib: Learning single-image calibration with geometric optimization},
  author       = {Veicht, Alexander and Sarlin, Paul-Edouard and Lindenberger, Philipp and Pollefeys, Marc},
  booktitle    = {European Conference on Computer Vision},
  pages        = {1--20},
  year         = {2024},
  organization = {Springer}
}

@inproceedings{zhang2025enabling,
  title={Enabling instructional image editing with in-context generation in large scale diffusion transformer},
  author={Zhang, Zechuan and Xie, Ji and Lu, Yu and Yang, Zongxin and Yang, Yi},
  booktitle={The Thirty-ninth Annual Conference on Neural Information Processing Systems},
  year={2025}
}

@article{liu2025step1x,
  title   = {Step1x-edit: A practical framework for general image editing},
  author  = {Liu, Shiyu and Han, Yucheng and Xing, Peng and Yin, Fukun and Wang, Rui and Cheng, Wei and Liao, Jiaqi and Wang, Yingming and Fu, Honghao and Han, Chunrui and others},
  journal = {arXiv preprint arXiv:2504.17761},
  year    = {2025}
}

@article{wu2025omnigen2,
  title   = {Omnigen2: Exploration to advanced multimodal generation},
  author  = {Wu, Chenyuan and Zheng, Pengfei and Yan, Ruiran and Xiao, Shitao and Luo, Xin and Wang, Yueze and Li, Wanli and Jiang, Xiyan and Liu, Yexin and Zhou, Junjie and others},
  journal = {arXiv preprint arXiv:2506.18871},
  year    = {2025}
}

@misc{flux2dev,
  author       = {{Black Forest Labs}},
  title        = {FLUX.2-dev},
  year         = {2025},
  publisher    = {Hugging Face},
  howpublished = {\url{https://huggingface.co/black-forest-labs/FLUX.2-dev}}
}

@article{cao2025hunyuanimage,
  title   = {Hunyuanimage 3.0 technical report},
  author  = {Cao, Siyu and Chen, Hangting and Chen, Peng and Cheng, Yiji and Cui, Yutao and Deng, Xinchi and Dong, Ying and Gong, Kipper and Gu, Tianpeng and Gu, Xiusen and others},
  journal = {arXiv preprint arXiv:2509.23951},
  year    = {2025}
}

@misc{gptimage15,
  author       = {{OpenAI}},
  title        = {GPT Image 1.5 Model},
  year         = {2025},
  howpublished = {\url{https://developers.openai.com/api/docs/models/gpt-image-1.5}}
}

@misc{nanobananapro,
  author       = {Naina Raisinghani},
  title        = {Introducing {Nano Banana Pro}},
  year         = {2025},
  month        = {November},
  howpublished = {\url{https://blog.google/innovation-and-ai/products/nano-banana-pro/}},
  note         = {Google DeepMind Blog. Accessed: 2026-04-02}
}

@article{wan2025wan,
  title   = {Wan: Open and advanced large-scale video generative models},
  author  = {Wan, Team and Wang, Ang and Ai, Baole and Wen, Bin and Mao, Chaojie and Xie, Chen-Wei and Chen, Di and Yu, Feiwu and Zhao, Haiming and Yang, Jianxiao and others},
  journal = {arXiv preprint arXiv:2503.20314},
  year    = {2025}
}

@article{zhu2026gide,
  title   = {{GIDE}: Unlocking Diffusion {LLM}s for Precise Training-Free Image Editing},
  author  = {Zhu, Zifeng and Han, Jiaming and Zhao, Jiaxiang and Luo, Minnan and Yue, Xiangyu},
  journal = {arXiv preprint arXiv:2603.21176},
  year    = {2026},
  doi     = {10.48550/arXiv.2603.21176},
  url     = {https://arxiv.org/abs/2603.21176}
}

\clearpage

\appendix


\renewcommand{\thefigure}{S\arabic{figure}}
\renewcommand{\thetable}{S\arabic{table}}
\setcounter{figure}{0}
\setcounter{table}{0}

\section*{Supplementary Material}

\begin{table*}[t]
\centering
\caption{Comparison of our training dataset with existing related datasets. Our dataset uniquely provides sequential transition frames (CoF) combined with explicit camera parameter pairs and high scene diversity.}
\label{tab:train_dataset_comparison}
\resizebox{\textwidth}{!}{
\begin{tabular}{lccccccc}
\toprule
\textbf{Dataset} & \textbf{Task} & \textbf{Size} & \textbf{Content Diversity} 
& \textbf{Synthetic Content} & \textbf{Sequential Transitions} 
& \textbf{Intrinsics} & \textbf{Extrinsics} \\ 
\midrule
RealEstate10K \cite{he2024cameractrl} & Video Synthesis    & 10M frames   & Low          & No  & No         & vFoV, distortion       & roll, pitch, yaw \\
Puffin-4M \cite{liao2025thinking}     & Image Generation   & 4M triplets  & Medium--High & No  & No         & vFoV                   & roll, pitch, yaw \\
PreciseCam\cite{bernal2025precisecam} & T2I Generation & 57,380 imgs & Medium--High & No & No & vFoV, distortion & roll, pitch \\
CamEdit\cite{qin2025camedit} & Image Editing & 50K pairs  & High         & Yes & No         & aperture, focal length & focal plane \\
\midrule
\rowcolor{gray!10}
\textbf{Ours} & \textbf{Image Editing} & \textbf{5760 seqs} & \textbf{High}
    & \textbf{Yes} & \textbf{Yes (CoF)} & \textbf{vFoV, radial dist} 
    & \textbf{roll, pitch, yaw} \\
\bottomrule
\end{tabular}
}
\end{table*}

\begin{figure*}[t] 
    \centering
    \includegraphics[width=\textwidth]{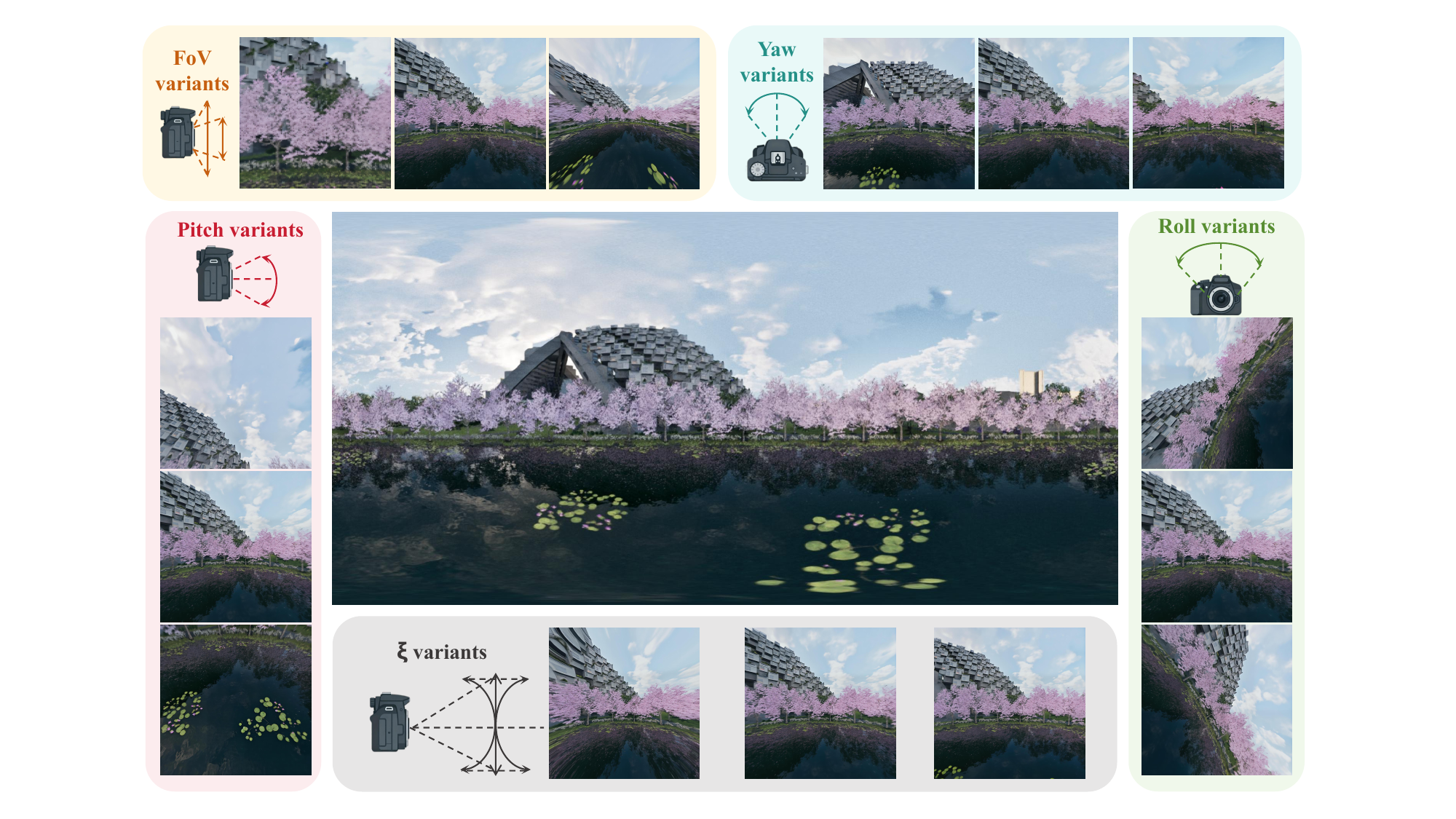} 
    \caption{Visual impact of individual camera parameters on the cropped perspective. We systematically manipulate yaw, pitch, roll, vertical FOV, and radial distortion ($\xi$) to demonstrate their distinct geometric transformations on the base panorama.}
    \Description{Examples showing how yaw, pitch, roll, vertical field of view, and radial distortion change a cropped panorama.}
    \label{fig:camera_params}
\end{figure*}

This supplementary material provides additional implementation details, extended experimental analysis, and qualitative results to support the main manuscript. The content is organized as follows:

\begin{itemize}
    \item \textbf{Automated Pipeline for Dataset Construction (Appendix A):} Details the dataset parameter sampling strategy, temporal sequence interpolation, VLM filtering prompts, and comparisons with existing datasets.
    
    \item \textbf{CamEditor-Bench and Evaluation Protocols (Appendix B):} Provides the mathematical derivation of the Target Perspective Field (PF-T) and a comprehensive comparison of our benchmark against existing ones.
    
    \item \textbf{Additional Experimental Details (Appendix C):} Explains the deterministic instruction prompt conversion for baseline models and discusses the quantitative impact of perception module accuracy.
    
    \item \textbf{Visualization (Appendix D):} Presents further qualitative visualization results and comparative generation examples.
    
    \item \textbf{Limitations (Appendix E):} Discusses the physical and algorithmic bounds of our framework, including resolution trade-offs and metric estimation noise.
\end{itemize}

\section{Automated Pipeline for Dataset Construction}

\subsection{Training Dataset Comparison and Necessity}

To train a model capable of precise camera-parameterized editing via sequential modeling, the training data must simultaneously satisfy three critical criteria: explicit mathematical camera annotations, continuous temporal transitions, and high content diversity. As summarized in Table \ref{tab:train_dataset_comparison}, existing datasets fail to meet all these requirements simultaneously, necessitating the construction of our specialized dataset.

Specifically, traditional video datasets like RealEstate10K \cite{he2024cameractrl} provide rich camera trajectories but lack paired editing images, and their content diversity is strictly limited to indoor and outdoor property layouts. Recent camera-aware datasets, including Puffin-4M \cite{liao2025thinking}, PreciseCam\cite{bernal2025precisecam}, and CamEdit\cite{qin2025camedit}, introduce physical camera parameters. However, they are inherently designed for single-step generation or static image pairs, lacking the intermediate sequential transitions.

In contrast, our dataset leverages a hybrid source of real-world panoramas and precise Unreal Engine 5 (UE5) synthetic environments, achieving high content diversity across complex indoor layouts, urban streets, natural landscapes, and dynamic human activities. It provides 5,760 high-quality sequences with absolute mathematical mappings (vFoV, radial distortion, roll, pitch, yaw). Crucially, it is the only dataset incorporating Chain of Frames (CoF) sequential transitions. This unique temporal structure empowers our video-prior framework to effectively decompose large-magnitude perspective shifts and prevent structural tearing.

\subsection{Dataset Parameter Sampling Strategy}

We model the camera behavior using a unified perspective field representation parameterized by five variables: yaw, pitch, roll, vertical Field of View (vFoV), and radial distortion ($\xi$) \cite{jin2023perspective}. The visual impact of each individual parameter is illustrated in Figure~\ref{fig:camera_params}.

To ensure a comprehensive and balanced distribution of camera parameters across our training dataset, we employ a rigorous stratified grid sampling strategy. The generation pipeline is governed by four core perspective parameters: roll $r \in [-90^\circ, 90^\circ]$, pitch $p \in [-90^\circ, 90^\circ]$, vertical Field of View (vFoV) $v \in [30^\circ, 140^\circ]$, and radial distortion $\xi \in [0, 1]$. 

To cover this high-dimensional continuous space and prevent sampling bias towards common viewpoints, we discretize the parameter ranges into uniformly distributed bins. Specifically, the parameter space is partitioned into 6 bins for roll, 6 bins for pitch, 4 bins for vFoV, and 4 bins for radial distortion. This orthogonal division yields a total of $6 \times 6 \times 4 \times 4 = 576$ distinct parameter subspace grids.

Within each of the 576 grid combinations, we randomly sample exactly 10 specific parameter sets. This strategic design ensures both macro-level structural diversity across the entire geometric domain and micro-level stochasticity within local parameter neighborhoods, deterministically generating the final 5,760 strictly aligned reference-target training instances utilized in our framework.

\subsection{Parameter Pairing and Sequence Interpolation}

For each sampled base configuration $p_{base}$, we generate a corresponding target configuration $p_{edited}$. To ensure the resulting image pairs maintain strict structural correspondence and \textbf{remain strictly within the bounds of controllable editing rather than unconstrained outpainting}, we impose conditional bounds on $p_{edited}$. Specifically, while yaw is strictly locked, pitch and FOV are constrained within a localized operational window relative to the base state:

\[
\begin{cases}
    \text{yaw}_{edited} = \text{yaw}_{base} \\
    \text{pitch}_{edited} \sim U(\text{pitch}_{base} - 40^\circ, \text{pitch}_{base} + 40^\circ) \\
    \text{roll}_{edited} \sim U(-90^\circ, 90^\circ) \\
    \text{fov}_{edited} \sim U(\text{fov}_{base} - 40^\circ, \text{fov}_{base} + 40^\circ) \\
    \xi_{edited} \sim U(0.0, 1.0)
\end{cases}
\]

Using these constrained parameter pairs, we execute a spherical projection to crop the base and target images directly from the high-resolution panoramas, establishing perfectly aligned reference-target pairs.

To construct the intermediate transition sequence necessary for our Chain of Frames (CoF) strategy, we linearly interpolate the camera parameters between the base and edited states. The parameter configuration $p^{i}$ for the $i$-th intermediate frame ($i \in \{1, 2, \dots, N\}$) is mathematically defined as:

\[
p^{i} = (1 - \frac{i}{N+1}) p_{base} + \frac{i}{N+1} p_{edited}, \quad i \in \{1,2, \dots, N\}.
\] 

This interpolation guarantees a mathematically continuous geometric trajectory across the temporal sequence, providing the video diffusion model with a coherent structural prior.

\subsection{VLM Prompt for Training Data Filtering}

As detailed in Section 4.1 (Stage 3) of the main text, we utilize a Vision-Language Model (VLM) as an automated data cleaning mechanism to filter the synthesized training instances. 

To ensure the full reproducibility of our data generation pipeline, the evaluation used the open-weights \texttt{Qwen2.5-VL-7B-Instruct} model released by Qwen. The input provided to the model is a single wide image, constructed by horizontally concatenating the unedited base image and its corresponding edited output. To ensure deterministic scoring and strictly eliminate generation randomness, the decoding temperature of the model was set to 0.0. 

The VLM acts as a rigorous data annotator, scoring each cropped base-edited pair on a scale of 0 to 10. As defined in our pipeline, only image pairs that maintain a strict minimum score threshold across all five specified dimensions are retained for the final training set. The exact system prompt provided to the model is presented below.

\begin{tcolorbox}[colback=gray!5!white,colframe=black!75!black,title=System Prompt for Training Data Filtration, breakable]
\small
\textbf{System Role:} You are an expert AI data evaluator, specializing in assessing the quality of image pairs for training a novel view synthesis model. Your task is to analyze a given image and provide a quality score using a deduction system.

The input image is a single wide image, which is a horizontal concatenation of two different views of the same 3D scene ("View A" and "View B"). The placement of these views (left or right) is random.

\textbf{Scoring Instructions:}
\begin{itemize}
    \item The image pair starts with a full score of 10.
    \item For each criterion below, assign a sub-score between 0 (completely unsatisfactory) and 10 (fully satisfactory), and briefly explain your reasoning.
    \item The final score is the \textbf{lowest} of all sub-scores (not the average). This means that a low score in any single aspect will significantly lower the overall score, ensuring that only images that are strong in all aspects receive a high final score.
    \item If a criterion is severely violated, give a low sub-score (e.g., 0\textasciitilde 3); if only minor issues exist, give a moderate sub-score (e.g., 4\textasciitilde 7).
    \item Use the full range of scores to reflect subtle differences.
\end{itemize}

\textbf{Criteria (score each separately):}

\textbf{1. Correlation and Learnability:} 
Are there clear, corresponding features or objects between the two views, making the geometric transformation learnable? \textit{Do not penalize for drastic perspective changes or non-linear distortions if correlation is maintained.}
\begin{itemize}
    \item "Corresponding features or objects" means that most major objects or regions in one view can be clearly matched to those in the other view, even if their positions or shapes change due to perspective.
    \item It is acceptable for a small number of objects to be missing or not fully corresponding due to large viewpoint changes, occlusion, or field-of-view differences. Do not penalize heavily for such cases as long as the majority of core subjects and the overall spatial structure remain consistent and learnable.
    \item If most major objects can be matched and the spatial structure is reasonable, give a high score (8-10). If some objects are missing but the core content still corresponds, give a moderate score (4-7). Only if most content does not correspond or the spatial relationship is chaotic, give a low score (0-3).
    \item When giving a score, briefly state the main reason for any deduction.
\end{itemize}

\textbf{2. Information Balance:} 
Do both views contain a comparable amount and complexity of information regarding the core subjects? \textit{Major imbalance (one view rich, one view sparse) should be heavily penalized.}
\begin{itemize}
    \item "Main subject" refers to an object or region in the image with clear boundaries and independent semantic meaning. Large areas of ground, sky, or wall usually count as only one subject.
    \item If there is a drastic drop in the number of main subjects between the two views (e.g., 10 in the left and only 1 in the right), even if the area is similar, the score should be very low (e.g., 1-2).
    \item Only when the number of main subjects is similar in both views should you further consider the area, texture complexity, and other details to fine-tune the score.
    \item Please make full use of the 0-10 score range, and do not only give middle scores. When there is a significant difference in subject number, be bold in giving low scores.
\end{itemize}

\textbf{3. Subject Richness and Scene Complexity:} 
Does each view contain rich and varied content---multiple meaningful subjects, some interactions, or a single subject with rich structure/texture? Evaluate left and right views separately; use the lower score.
\begin{itemize}
    \item \textbf{High score (8--10):} The view contains several distinct subjects, or a single subject with rich detail/structure. As long as the scene is not monotonous, give a high score.
    \item \textbf{Medium score (5--7):} The view has some content, but is not particularly rich or diverse. 
    \item \textbf{Low score (0--4):} The view is extremely simple or monotonous (e.g., only ground, sky, or a blank wall).
    \item Do not be too conservative---give high scores to any scene that is clearly not monotonous, regardless of the scene category.
\end{itemize}

\textbf{4. Image Clarity:} 
Is the image reasonably clear? \textit{Evaluate the left and right images separately, and use the lower of the two scores.} Minor blur is acceptable, but if there are large areas (e.g., more than 20\% of the image) where all texture and detail are lost due to severe blur, haze, or overexposure, this should be heavily penalized.

\textbf{5. Content Coherence \& Artifacts:} 
Is the image free of major non-geometric distractions (e.g., illogical breaks at the seam, prominent watermarks, or large irrelevant occluders)? \textit{Evaluate the left and right images separately, and use the lower of the two scores.} Apply a minor penalty for such issues.

\textbf{Output Format:} 
You MUST provide your response in the following strict format:

Sub-scores: \\
Correlation and Learnability: [0\textasciitilde 10] Reason: [...] \\
Information Balance: [0\textasciitilde 10] Reason: [...] \\
Subject Richness and Scene Complexity: [0\textasciitilde 10] Reason: [...] \\
Image Clarity: [0\textasciitilde 10] Reason: [...] \\
Content Coherence \& Artifacts: [0\textasciitilde 10] Reason: [...]

Final Score: [This MUST be exactly the minimum of all sub-scores above. Do NOT use average, weighted sum, or any other method. There must be no exceptions.] \\
Summary: [A concise, one-to-two-sentence explanation focusing on the most influential positive and negative factors.]

\textbf{Important Enforcement:}
\begin{itemize}
    \item You MUST first list all sub-scores, then explicitly identify the minimum value among them.
    \item The Final Score MUST be exactly this minimum value, with no exceptions.
    \item Example of a common mistake: If your sub-scores are 6, 8, 4, 9, 9, the Final Score MUST be 4. If you output 6, this is WRONG.
\end{itemize}
\end{tcolorbox}

\section{CamEditor-Bench and Evaluation Protocols}

\subsection{Comprehensive Benchmark Comparison: Data Properties and Evaluation Metrics}

To highlight the unique positioning and necessity of CamEditor-Bench, Table \ref{tab:benchmark_comparison_details} systematically compares it with three representative benchmarks across two core aspects: \textit{Data Properties} and \textit{Evaluation Metrics}.

\textbf{1. Superiority in Data Properties:}\\
Existing benchmarks are sub-optimal for precise camera editing. Video datasets such as RealEstate10K address synthesis rather than editing and lack paired instructions. Recent camera-centric benchmarks (PreciseCam, Puffin-Gen \cite{liao2025thinking}) focus on pure text-to-image (T2I) generation and omit the structural constraints of source-conditioned editing. CamEditor-Bench instead targets image editing. By adopting a \textit{Source Image + Params} modality, it completely bypasses textual spatial ambiguity and provides an absolute mathematical transformation target.

\textbf{2. Rigor in Evaluation Metrics:}\\
Traditional metrics fail to decouple content preservation from massive perspective shifts. While PreciseCam and Puffin-Gen utilize optimization-free geometric representations for T2I generation, they ignore the source identity preservation dilemma inherent in editing tasks. To address this, CamEditor-Bench introduces a fully decoupled evaluation protocol. We combine DINO-v2, CLIP-I, SSIM, and LPIPS to rigorously monitor semantic and structural fidelity. Concurrently, we pioneer Target-Condition (PF-T) and GT-Image (PF-I) alignments based on Perspective Fields ($S_{up}$, $E_{lat}$). This provides an absolute, optimization-free quantification of geometric precision, preventing the estimator-collapse issues common in traditional SfM-based metrics. The rigorous mathematical derivation of these geometric metrics is detailed in the following subsection.

\begin{table*}[t]
\centering
\caption{Comprehensive comparison of CamEditor-Bench with representative benchmarks. The comparison is strictly categorized into two primary aspects: \textit{Data Properties} and \textit{Evaluation Metrics}. Our benchmark uniquely evaluates precise camera editing using a Source Image + Params modality, combined with a fully decoupled, optimization-free geometric evaluation protocol.}
\label{tab:benchmark_comparison_details}
\resizebox{\textwidth}{!}{
\begin{tabular}{lcccccc}
\toprule
\multirow{2}{*}{\textbf{Benchmark}} & \multicolumn{4}{c}{\textbf{Data Properties}} & \multicolumn{2}{c}{\textbf{Evaluation Metrics}} \\
\cmidrule(lr){2-5} \cmidrule(lr){6-7}
& \textbf{Test Size} & \textbf{Core Task Focus} & \textbf{Data Source} & \textbf{Instruction Modality} & \textbf{Content \& Quality} & \textbf{Geometry \& Camera (Opt-Free)} \\ 
\midrule
RealEstate10K-test \cite{he2024cameractrl} & 7,289 seqs       & Video Synthesis          & Real YouTube Videos  & Camera Trajectory             & PSNR, SSIM, LPIPS           & N/A \\
PreciseCam \cite{bernal2025precisecam}         & 57,380$^\dagger$ & Camera-guided T2I Gen    & Real Panoramas       & Text + Params                 & CLIPScore                   & PF-US ($u_x$, $\phi_x$) \\
Puffin-Gen \cite{liao2025thinking}        & 650              & Unified T2I Gen \& Und   & Real Panoramas + Syn & Text + Params                 & N/A                         & Up-vector, Latitude, Gravity \\
\midrule
\rowcolor{gray!10}
\textbf{Ours (CamEditor-Bench)} & \textbf{462} & \textbf{Camera-Parameterized Edit} & \textbf{Real Panoramas + UE5} & \textbf{Source Image + Params} & \textbf{DINO-v2, CLIP-I, SSIM, LPIPS} & \textbf{PF-T, PF-I ($S_{up}$, $E_{lat}$)} \\
\bottomrule
\end{tabular}
}
\vspace{0.5em}
{\small $^\dagger$Full dataset size; no independent held-out evaluation split is defined.}
\end{table*}

\subsection{Derivation of the Target Perspective Field (PF-T)}

In Section 4.3 of the main text, we introduced the PF-T metric to evaluate the absolute geometric alignment of the generated image against the user-provided target camera conditions. Unlike traditional parameter estimation methods that rely on non-linear optimization and suffer from inherent domain biases, PF-T provides an optimization-free evaluation by computing dense geometric representations directly from the given parameter set.

Given a user-specified camera configuration
\[
\Omega = \{\text{roll}, \text{pitch}, \text{vFoV}, \xi\},
\]
our goal is to analytically derive the target Perspective Field, which consists of a dense latitude map $\varphi$ and an up-vector field $\mathbf{u}$ for every pixel $\mathbf{x} = (u, v)$.

\textbf{1. Projection Model Formulation.} We employ the Unified Spherical (US) camera model to map any 3D spatial point $\mathbf{X} = (x, y, z)$ to the 2D image plane. This model effectively accommodates complex non-linear distortions $\xi$ and large field-of-views. The projection function $\mathcal{P}(\mathbf{X}) = \mathbf{x}$ is defined as: 
$$
\mathcal{P}(\mathbf{X}) = \begin{bmatrix} u \\ v \end{bmatrix} = \begin{bmatrix} \frac{xf}{\xi \sqrt{x^2+y^2+z^2} + z} + u_0 \\ \frac{yf}{\xi \sqrt{x^2+y^2+z^2} + z} + v_0 \end{bmatrix}
$$
where $(u_0, v_0)$ represents the principal point, and the focal length $f$ is derived directly from the user-specified vFoV.

\textbf{2. Latitude Map Derivation}

The target extrinsic parameters (roll and pitch) determine the orientation of the camera relative to the world coordinate system, allowing us to establish the gravity vector $\mathbf{g}$ within the camera's reference frame. For any pixel $\mathbf{x}$, the corresponding 3D light ray emitted from the camera center is denoted as $\mathbf{R}$. The target latitude $\varphi_{\mathbf{x}}$, which measures the elevation angle between the light ray $\mathbf{R}$ and the horizontal world plane, is computed as: 
$$
\varphi_{\mathbf{x}} = \arcsin\left(\frac{\mathbf{R} \cdot \mathbf{g}}{\|\mathbf{R}\|_2}\right)
$$

\textbf{3. Up-Vector Field Derivation}

The target up-vector $\mathbf{u}_{\mathbf{x}}$ at pixel $\mathbf{x}$ represents the 2D image-plane projection of the absolute upward direction in the 3D world (which is strictly opposite to the gravity vector $\mathbf{g}$). Utilizing the projection function $\mathcal{P}$, the dense up-vector field is mathematically derived by computing the limit of the projection difference:
$$
\mathbf{u}_{\mathbf{x}} = \lim_{c \to 0} \frac{\mathcal{P}(\mathbf{X} - c\mathbf{g}) - \mathcal{P}(\mathbf{X})}{\|\mathcal{P}(\mathbf{X} - c\mathbf{g}) - \mathcal{P}(\mathbf{X})\|_2}
$$
By generating the target latitude map $\varphi_T$ and up-vector field $\mathbf{u}_T$ through this deterministic formulation, we establish a mathematically rigorous anchor. As described in the main text, the absolute geometric precision of the generative models is subsequently quantified by computing the Mean Squared Error (MSE) of the latitude values $E_{lat}^T$ and the cosine similarity of the upward vectors $S_{up}^T$.

\section{Additional Experimental Details}

\subsection{Baseline Input Parity and Selection}

Table~\ref{tab:baseline_inputs} summarizes the information available to each
method. The two Wan2.1 controls and CameraEditor use the same
Wan2.1-T2V-14B backbone, training data, and optimization settings. The
no-reference control receives the target camera through the same deterministic
prompt template as the instruction-driven editors. The Pl\"ucker control
instead receives explicit camera embeddings derived from GeoCalib. Both
controls retain CoF, so the comparison isolates the effect of the conditioning
representation and visual reference sequence.

\begin{table*}[t]
\centering
\caption{Input parity and backbone configuration for the evaluated method
families. ``Det. prompt'' denotes the deterministic template in
Table~\ref{tab:prompt_mapping}.}
\label{tab:baseline_inputs}
\begin{tabular}{lcccc}
\toprule
\textbf{Method family} & \textbf{Camera signal} & \textbf{GeoCalib} &
\textbf{Reference sequence} & \textbf{Backbone} \\
\midrule
Open-source editors & Det. prompt & No & No & Various \\
Closed APIs & Det. prompt & No & No & Undisclosed \\
Wan2.1, no reference & Det. prompt & No & No & Wan2.1-T2V-14B \\
Wan2.1 + Pl\"ucker & Pl\"ucker embedding & Yes & No & Wan2.1-T2V-14B \\
\rowcolor{gray!10}
\textbf{CameraEditor} & \textbf{Camera parameters} & \textbf{Yes} &
\textbf{Yes} & \textbf{Wan2.1-T2V-14B} \\
\bottomrule
\end{tabular}
\end{table*}

We do not include CamEdit \cite{qin2025camedit} as a numerical baseline
because its released task definition controls a disjoint parameter space
(aperture, focal length, and focal plane), and no public implementation was
available at the time of evaluation. A geometric warp followed by inpainting
cannot provide ground-truth-aligned content in newly exposed regions, while
direct panorama reprojection requires a panoramic user input. We therefore
report these approaches as task-boundary references rather than numerical
comparators for single-perspective-image editing.

\subsection{Instruction Prompt Conversion for Baselines}

Existing instruction-driven models struggle to interpret continuous numerical parameters. To ensure a strictly fair comparison, we establish a deterministic mapping (Table~\ref{tab:prompt_mapping}) that translates our exact target camera configurations ($p_{edited}$) into hybrid natural language prompts comprehensible to baseline models.

Specifically, for extrinsics ($roll$, $pitch$) and vertical Field of View ($vFoV$), we partition their mathematical domains into standard cinematographic terminology while dynamically injecting the exact numerical values to avoid quantization penalties. Since standard text-to-image baselines are agnostic to radial distortion ($\xi$), we map its coefficient range $(0, 1)$ into explicit optical lens effects (e.g., from "rectilinear lens" to "extreme spherical fisheye").

\begin{table}[htbp]
\centering
\caption{Deterministic mapping of continuous camera parameters to hybrid natural language templates for baseline evaluation. The exact numerical parameter (e.g., $roll, pitch$) is dynamically injected into the text to avoid quantization penalties.}
\label{tab:prompt_mapping}
\resizebox{\linewidth}{!}{
\begin{tabular}{lll}
\toprule
\textbf{Parameter} & \textbf{Value Range ($p$)} & \textbf{Dynamic Natural Language Template ($t$)} \\
\midrule
\multirow{5}{*}{\textbf{Roll}} 
& $[-90^\circ, -45^\circ)$ & Extreme counterclockwise Dutch angle of $roll^\circ$ \\
& $[-45^\circ, -10^\circ)$ & Counterclockwise Dutch angle of $roll^\circ$ \\
& $[-10^\circ, 10^\circ]$ & Near level shot of $roll^\circ$ \\
& $(10^\circ, 45^\circ]$ & Clockwise Dutch angle of $roll^\circ$ \\
& $(45^\circ, 90^\circ]$ & Extreme clockwise Dutch angle of $roll^\circ$ \\
\midrule
\multirow{5}{*}{\textbf{Pitch}} 
& $[-90^\circ, -45^\circ)$ & Extreme high-angle (Bird's-eye view) of $pitch^\circ$ \\
& $[-45^\circ, -10^\circ)$ & High-angle shot (Tilt-down) of $pitch^\circ$ \\
& $[-10^\circ, 10^\circ]$ & Near straight-on shot of $pitch^\circ$ \\
& $(10^\circ, 45^\circ]$ & Low-angle shot (Tilt-up) of $pitch^\circ$ \\
& $(45^\circ, 90^\circ]$ & Extreme low-angle (Worm's-eye view) of $pitch^\circ$ \\
\midrule
\multirow{4}{*}{\textbf{vFoV}} 
& $[30^\circ, 50^\circ)$ & Close-up shot with a $vFoV^\circ$ vertical field of view \\
& $[50^\circ, 75^\circ)$ & Medium shot with a $vFoV^\circ$ vertical field of view \\
& $[75^\circ, 100^\circ)$ & Wide-angle shot with a $vFoV^\circ$ vertical field of view \\
& $[100^\circ, 140^\circ]$ & Ultra wide-angle shot with a $vFoV^\circ$ vertical field of view \\
\midrule
\multirow{4}{*}{\textbf{Distortion ($\xi$)}} 
& $[0, 0.2)$ & Standard rectilinear lens with a distortion coefficient of $\xi$ \\
& $[0.2, 0.5)$ & Mild barrel distortion with a coefficient of $\xi$ \\
& $[0.5, 0.8)$ & Pronounced fisheye effect with a coefficient of $\xi$ \\
& $[0.8, 1.0]$ & Extreme spherical fisheye with a coefficient of $\xi$ \\
\bottomrule
\end{tabular}
}
\end{table}

A critical vulnerability in prompting generic editing models is "concept bleeding," where geometric instructions are overshadowed by original semantic descriptions. To mitigate this, we design a unified instruction template that strictly decouples structural camera modifiers from the core semantic content. For an instance with scene description $S_{desc}$ and mapped camera terms ($t_{roll}, t_{pitch}, t_{vFoV}, t_{\xi}$), the prompt is constructed as follows:

\begin{tcolorbox}[colback=gray!5!white,colframe=black!75!black,title=Unified Baseline Prompt Architecture, breakable]
\small
\textbf{Task Instruction:} \\
"Edit the provided image to match the specified camera perspective and optical lens effects, while strictly maintaining the original scene identity and structural integrity."

\textbf{Camera Modifiers:} \\
"Perspective: Apply a \textbf{[ $t_{pitch}$ ]} and a \textbf{[ $t_{roll}$ ]}." \\
"Lens Effect: Use a \textbf{[ $t_{vFoV}$ ]} combined with a \textbf{[ $t_{\xi}$ ]}."

\textbf{Semantic Content:} \\
"Original Scene: \textbf{[ $S_{desc}$ ]}"
\end{tcolorbox}

By standardizing this architecture, we isolate the baselines' intrinsic geometric alignment capacities from variations in manual prompt engineering. This ensures that spatial coherence failures stem directly from architectural limitations.

\subsection{Human Evaluation Protocol}

The human study contains 300 blind pairwise comparisons: 100 against
Step1X-Edit, 100 against Qwen-Image-Editing, and 100 against
Nano-Banana-Pro. Each comparison shows the source image, ground truth,
target-camera visualization, and two anonymous edited outputs. Evaluators
judge the overall better result by jointly considering perceived camera
correctness, source-content preservation, and artifact severity. A tie
contributes 0.5 preference to each method. The resulting baseline-win,
tie, and CameraEditor-win counts are reported in
Table~\ref{tab:human_study}.

\subsection{Detailed Configurations of Perception Modules}

To answer RQ2 (Section~\ref{sec:rq2}), we evaluate how initial camera perception accuracy ($\Omega_{est} = \{roll, pitch, vFoV, \xi\}$) impacts final generation quality. Table~\ref{tab:perception_errors} quantifies the estimation errors of the four evaluated strategies.

\textbf{1. Random Routing:} Uniformly samples parameters without visual perception. The catastrophic geometric misalignment (e.g., $E_{pitch}=54.00^\circ$) causes severe structural collapse during generation (DINO-v2: $0.6301$, Table~\ref{tab:perception_ablation}).

\textbf{2. Puffin (Base \& Thinking) \cite{liao2025thinking}:} We utilize the official automated pipeline (\texttt{understanding.py}) to regress parameters directly from the source image. While the Chain-of-Thought reasoning in Puffin (Thinking) slightly reduces $E_{roll}$ and $E_{vFoV}$, both variants yield an identical $E_{\xi}$ ($0.338$) to the Random baseline. This exposes a fundamental limitation: VLM-based models are blind to continuous non-linear radial distortions, inherently capping their absolute geometric alignment capabilities ($S_{up}^T \le 0.5752$).

\textbf{3. GeoCalib (Ours):} Our pipeline uses GeoCalib's geometric calibration optimizer to extract the dense Perspective Field. Specifically designed for non-linear lens physics, it drastically reduces the distortion error ($E_{\xi}=0.201$) and pitch error ($E_{pitch}=6.05^\circ$). This robust geometric anchor directly translates to our state-of-the-art generation performance, achieving the highest structural alignment ($S_{up}^T: 0.6089$) and content preservation (DINO-v2: $0.8569$). This confirms that precise analytical perception is crucial for coherent camera-parameterized editing.

\begin{table}[ht]
\centering
\caption{Parameter estimation Mean Absolute Error (MAE). $E_{roll}, E_{pitch}$, and $E_{vFoV}$ denote the angular errors, and $E_{\xi}$ denotes the distortion coefficient error.}
\label{tab:perception_errors}
\resizebox{\linewidth}{!}{
\begin{tabular}{lcccc}
\toprule
\textbf{Strategy} & \textbf{$E_{roll}$ ($\downarrow$)} & \textbf{$E_{pitch}$ ($\downarrow$)} & \textbf{$E_{vFoV}$ ($\downarrow$)} & \textbf{$E_{\xi}$ ($\downarrow$)} \\
\midrule
Random & $48.42^\circ$ & $54.00^\circ$ & $28.51^\circ$ & 0.338 \\
Puffin (Base) \cite{liao2025thinking} & $26.40^\circ$ & $12.07^\circ$ & $18.55^\circ$ & 0.338 \\
Puffin (Thinking) \cite{liao2025thinking} & $25.83^\circ$ & $12.10^\circ$ & $\mathbf{17.52^\circ}$ & 0.338 \\
\rowcolor{gray!10} \textbf{GeoCalib (Ours)} \cite{veicht2024geocalib} & $\mathbf{20.76^\circ}$ & $\mathbf{6.05^\circ}$ & $21.82^\circ$ & \textbf{0.201} \\
\bottomrule
\end{tabular}
}
\end{table}

\section{Visualization}

\begin{figure*}[htbp] 
    \centering
    \includegraphics[width=\textwidth]{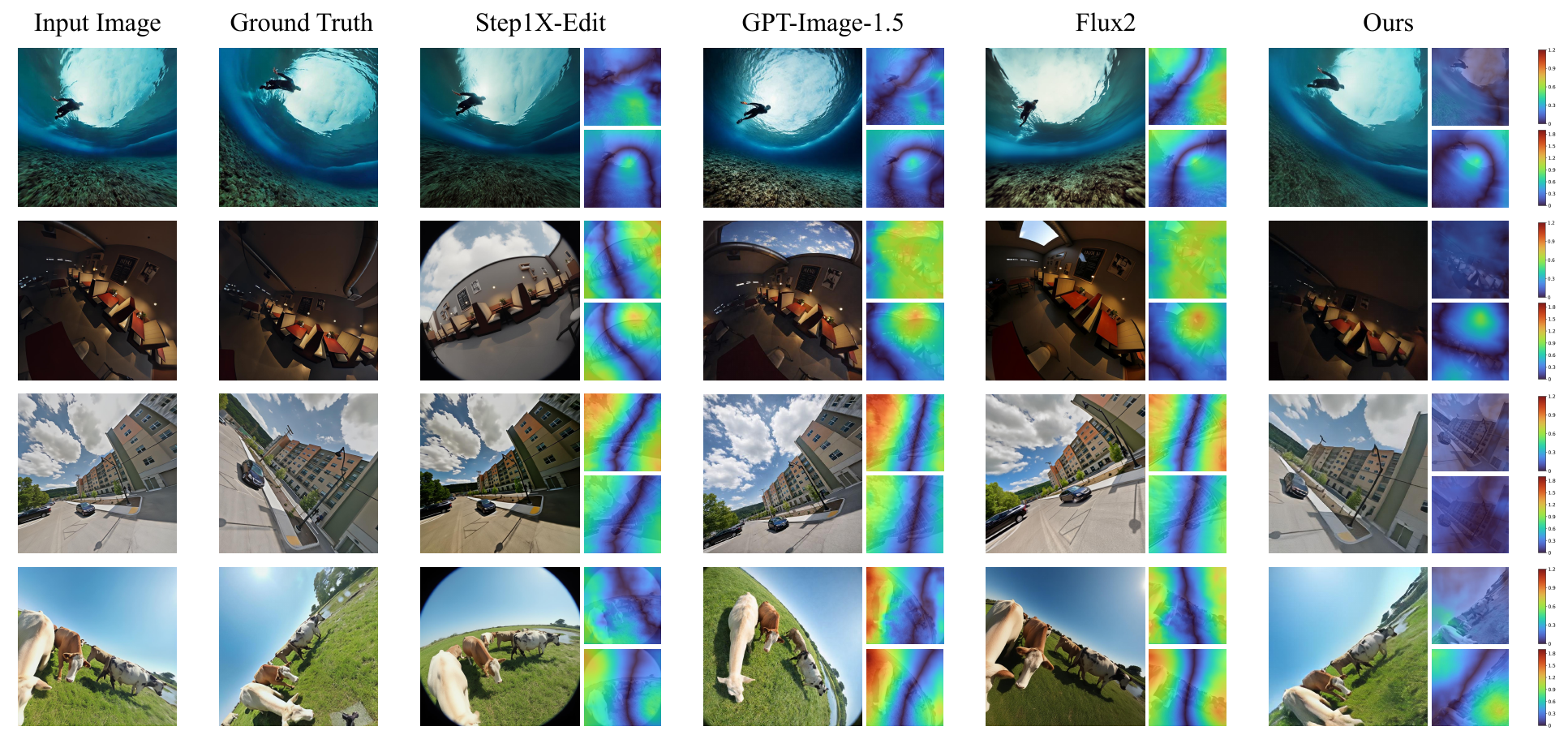} 
    \caption{Extended qualitative comparison against additional state-of-the-art editing models. Our framework accurately aligns with the target camera geometry---as verified by the corresponding Perspective Field (PF) heatmaps (right columns)---while strictly preventing the identity drift and structural hallucinations commonly observed in standard diffusion baselines.}
    \Description{Extended qualitative comparisons with perspective-field heatmaps for CameraEditor and competing editing methods.}
    \label{fig:main_comparison}
\end{figure*}

\begin{figure*}[htbp] 
    \centering
    \includegraphics[width=\textwidth]{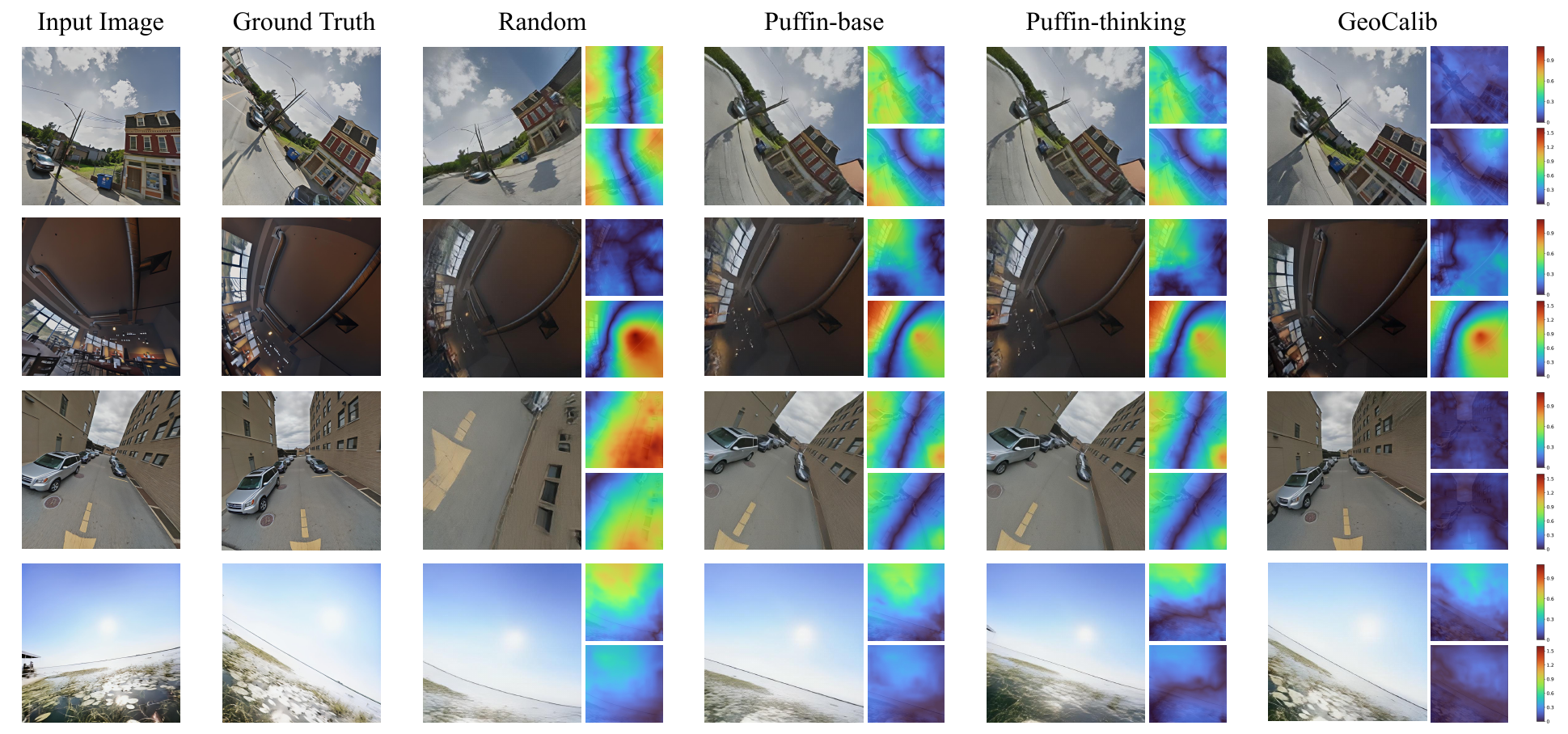} 
    \caption{Visual impact of different camera perception modules on the final generation quality. Erroneous initial estimations from Random routing or VLM-based understanding (Puffin variants) lead to catastrophic structural distortion. Our GeoCalib-driven routing provides a robust geometric anchor, ensuring physically coherent perspective editing.}
    \Description{Qualitative comparison of random, VLM-based, and GeoCalib-based camera perception modules.}
    \label{fig:perception_comparison}
\end{figure*}

\begin{figure*}[htbp] 
    \centering
    \includegraphics[width=\textwidth]{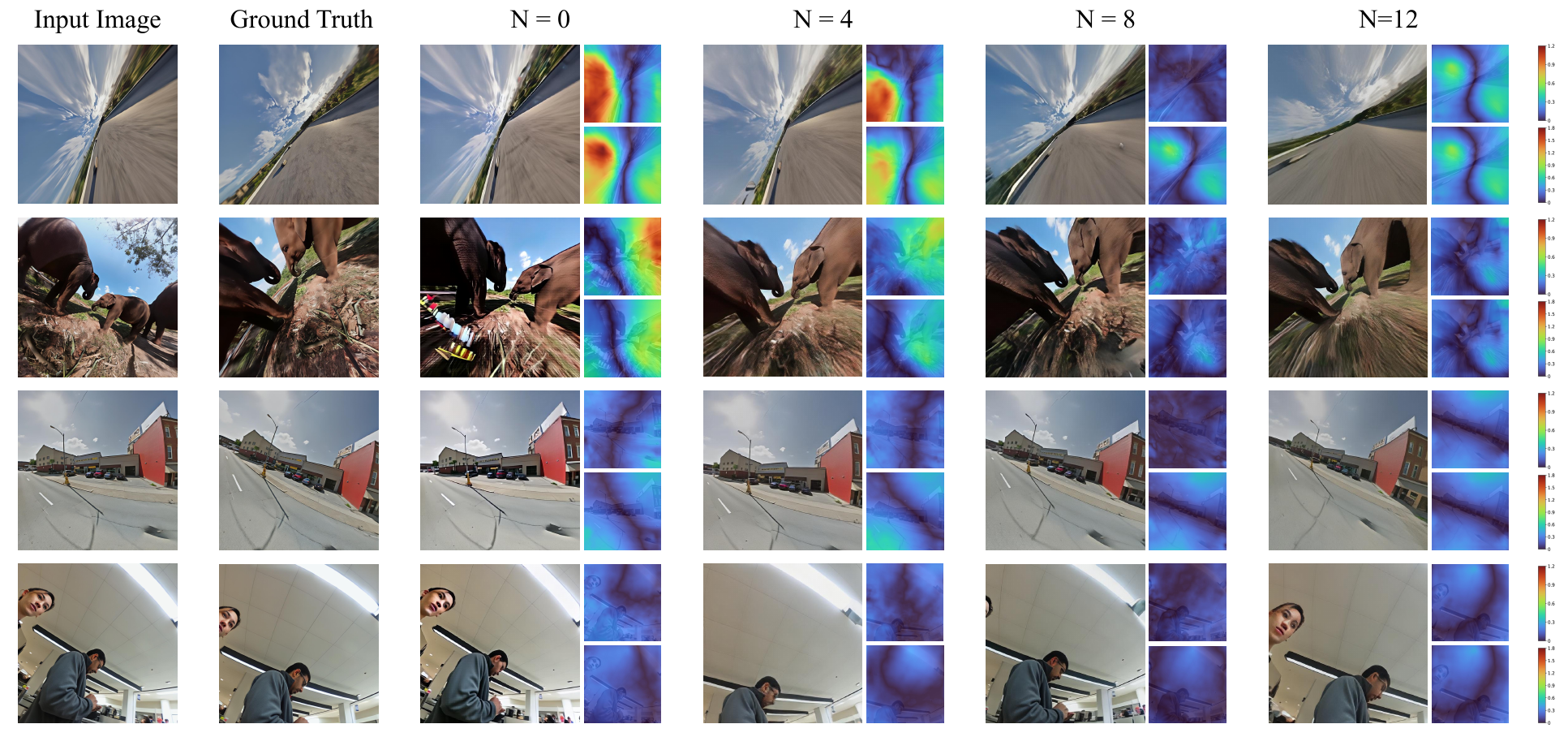} 
    \caption{Qualitative ablation on the number of intermediate transition frames ($N$) in our Chain-of-Frames (CoF) pipeline. Direct single-step generation ($N=0$) results in spatial tearing and identity loss under large perspective shifts. Increasing the temporal sequence length ($N=4, 8, 12$) provides stronger continuous geometric priors, significantly enhancing structural preservation.}
    \Description{Ablation examples for zero, four, eight, and twelve intermediate transition frames.}
    \label{fig:frames_ablation}
\end{figure*}

To further substantiate the quantitative findings presented in the main manuscript, this section provides extended qualitative visualizations. We systematically analyze the generative superiority of our framework, the critical role of accurate camera perception, and the structural impact of our Chain-of-Frames (CoF) temporal priors.

\subsection{Extended Baseline Comparisons}
Due to space constraints in the main text, Figure~\ref{fig:main_comparison} presents an extended qualitative comparison against additional state-of-the-art multi-modal and generative editing models, including Step1X-Edit, GPT-Image-1.5, and Flux2. While these models excel at semantic modifications, they fundamentally struggle with absolute spatial conditioning. Under large perspective shifts, they suffer from severe identity drift (hallucinating new structures) and fail to accurately match the target Perspective Field (PF), as evidenced by the misaligned up-vector and latitude heatmaps. In contrast, our method strictly anchors the generative process to the mathematical constraints, ensuring precise geometric alignment while perfectly preserving the source scene identity.

\subsection{Qualitative Impact of Perception Modules}
Figure~\ref{fig:perception_comparison} visualizes the downstream generative consequences of different initial camera perception strategies ($\Omega_{est}$), directly supporting the quantitative results in Table~\ref{tab:perception_errors}. When the routing mechanism relies on Random sampling or VLM-based estimators (Puffin-base, Puffin-thinking), the resulting geometric mismatch forces the diffusion model to warp the image into highly unnatural and physically impossible perspectives. By integrating GeoCalib, our pipeline secures a mathematically rigorous geometric anchor, completely eliminating these structural collapse artifacts and yielding coherent, high-fidelity perspective edits.

\subsection{Ablation on Intermediate Frames (N)}
Figure~\ref{fig:frames_ablation} provides a qualitative ablation on the sequence length ($N$) within our CoF framework. Attempting to bridge a massive perspective gap in a single step ($N=0$, standard image editing) leads to severe spatial tearing and outpainting hallucinations, as the non-linear pixel displacement exceeds the model's structural capacity. By explicitly decomposing the transformation into a continuous sequence ($N=4, 8, 12$), the video diffusion prior enforces strong inter-frame geometric consistency, significantly improving the structural integrity of the final edited frame. However, this geometric robustness introduces a fundamental trade-off: as $N$ increases, the accumulated temporal smoothing inherent to the sequence generation progressively compromises high-frequency textural clarity, leading to a noticeable loss of fine-grained details.

\section{Limitations}

While our proposed Chain-of-Frames (CoF) framework introduces a novel paradigm for structurally coherent, camera-parameterized image editing, it is bounded by several algorithmic and physical limitations that warrant future investigation:

\begin{itemize}
    \item \textbf{Trade-off Between Spatial Coherence and Textural Clarity:} 
    While our CoF framework ensures absolute geometric stability across massive perspective shifts, the sequential generation paradigm inherently suffers from cumulative errors. The temporal smoothing applied at each intermediate frame compounds as sequence length increases, progressively degrading high-frequency textures and single-frame clarity. Future work will explore hybrid latent upscaling to mitigate this degradation.
    
    \item \textbf{Inherent Ambiguities in 2D-to-3D Camera Estimation:} 
    Extracting explicit 3D geometry from 2D images remains fundamentally ill-posed, introducing bottlenecks at both ends of our pipeline. On the \textit{input} side, wild scenes with weak vanishing-point cues, reflective surfaces, or non-rigid content can cause the initial parameter estimation ($\Omega_{est}$) to drift, misguiding the entire editing sequence. On the \textit{evaluation} side, downstream estimators used for our PF-T metric can be tricked by complex generation artifacts, occasionally conflating generative failures with metric noise. Developing robust, generation-aware geometric perceivers is crucial.
\end{itemize}

\end{document}